\documentclass[]{fairmeta}

\usepackage{textcomp}
\usepackage{microtype}
\usepackage{graphicx}
\usepackage{booktabs} %
\usepackage{multirow}
\usepackage{hyperref}

\usepackage{amsmath}
\usepackage{amssymb}
\usepackage{bbm}
\usepackage{mathtools}
\usepackage{amsthm}
\usepackage{graphicx}
\usepackage{subcaption}
\usepackage{xurl}      %

\usepackage{algorithm}
\usepackage{algpseudocode}

\theoremstyle{plain}

\theoremstyle{definition}

\theoremstyle{remark}

\usepackage[textsize=tiny]{todonotes}

\title{Fast Byte Latent Transformer}

\author[1,2]{Julie Kallini}
\author[1]{Artidoro Pagnoni}
\author[1,3]{Tomasz Limisiewicz}
\author[1]{Gargi Ghosh}
\author[1,3]{Luke Zettlemoyer}
\author[2]{Christopher Potts}
\author[1]{Xiaochuang Han}
\author[1]{Srinivasan Iyer}

\affiliation[1]{FAIR at Meta}
\affiliation[2]{Stanford University}
\affiliation[3]{University of Washington}

\abstract{
Recent byte-level language models (LMs) match the performance of token-level models without relying on subword vocabularies, yet their utility is limited by slow, byte-by-byte autoregressive generation. We address this bottleneck in the Byte Latent Transformer (BLT) through new training and generation techniques. First, we introduce \textbf{BLT Diffusion \mbox{(BLT-D)}}, a new model and our fastest BLT variant, trained with an auxiliary block-wise diffusion objective alongside the standard next-byte prediction loss. This enables an inference procedure that generates multiple bytes in parallel per decoding step, substantially reducing the number of forward passes required to generate a sequence. Second, we propose two extensions inspired by speculative decoding that trade some of this speed for higher generation quality: \textbf{BLT Self-speculation \mbox{(BLT-S)}}, in which BLT's local decoder continues generating past its normal patch boundaries to draft bytes, which are then verified with a single full-model forward pass; and \textbf{BLT Diffusion+Verification \mbox{(BLT-DV)}}, which augments BLT-D with an autoregressive verification step after diffusion-based generation. All methods may achieve an estimated memory-bandwidth cost over 50\% lower than BLT on generation tasks. Each approach offers its own unique advantages, together removing key barriers to the practical use of byte-level LMs.
}

\date{\today}
\correspondence{Julie Kallini at \email{kallini@stanford.edu}, Srinivasan Iyer at \email{sviyer@meta.com}}

\begin{document}

\maketitle

\section{Introduction}
\label{sec:intro}

Byte-level (also known as \emph{tokenizer-free}) language models operate directly on raw bytes rather than a predefined vocabulary of tokens. By avoiding subword tokenization, they address several well-known shortcomings of token-level models, including sensitivity to input noise~\citep{pruthi-etal-2019-combating, sun2020adv0bert0}, handling structured or out-of-domain inputs~\citep{dagan-etal-2024-getting, singh2024tokenization, zhou-etal-2024-scaling}, limited character-level understanding~\citep{kaushal-mahowald-2022-tokens, huang-etal-2023-inducing, edman-etal-2024-cute}, and multilingual disparities~\citep{ahia-etal-2023-languages, petrov2023language, liang-etal-2023-xlm}. Despite their many advantages, byte-level models have seen limited adoption relative to subword models. The core issue is efficiency: since a typical subword token spans several bytes, a naively autoregressive byte-level model must operate over sequences that are many times longer than their token-level counterparts, dramatically increasing both training and inference cost~\citep{xue-etal-2022-byt5}.

Recent architectural innovations have substantially narrowed this efficiency gap. Rather than running a full Transformer over every byte, modern byte-level models often group bytes into larger units, use hierarchical computation, or replace full attention with more efficient sequence modeling mechanisms~\citep{el-boukkouri-etal-2020-characterbert, clark-etal-2022-canine, tay2022charformer, nawrot-etal-2022-hierarchical, nawrot-etal-2023-efficient, yu2023megabyte, slagle2024spacebyte, wang2024mambabyte, kallini2025mrt, evabyte, pagnoni-etal-2025-byte, hwang2025dynamicchunkingendtoendhierarchical}. For example, the \textbf{Byte Latent Transformer} (BLT; \citealt{pagnoni-etal-2025-byte}) dynamically groups bytes into variable-length \emph{patches} based on input complexity. Its hierarchical design concentrates computation on \emph{latent token} representations, allocating more compute to complex patches of text and yielding better scaling behavior than token-level models.

\begin{figure}[]
    \centering
    \includegraphics[height=0.35\textheight]{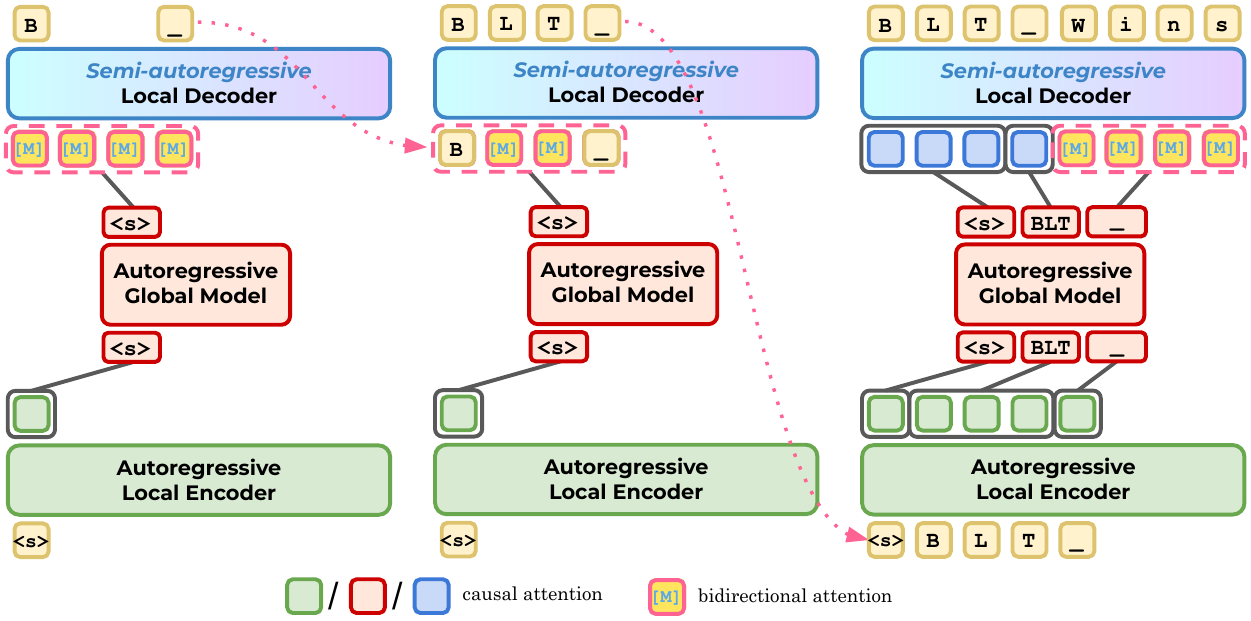}
    \caption{\textbf{BLT-D inference.}
    The encoder creates latent token representations from variable-length patches of bytes. The large global model predicts the next latent token. The decoder initializes a fixed-length block of $\mathtt{[MASK]}$ tokens and generates bytes in parallel via semi-autoregressive text diffusion, conditioning on the last latent token. \textbf{Compared to BLT, this inference approach decreases the forward passes/network function evaluations (NFEs) of all model components (encoder, global model, and decoder).}}
    \label{fig:inference}
\end{figure}

These advances reduce the \emph{compute} cost of byte-level models, but inference still faces a \emph{memory bandwidth} bottleneck. In modern LLM inference, generation cost is often dominated by repeatedly loading model weights and accessing key-value caches~\citep{pope-etal-2023-efficiently, kwon-etal-2023-efficient, yuan2024llminferenceunveiledsurvey}. Even when most computation is performed over latent token representations, standard byte-level decoding still generates one byte at a time. Since a typical subword token corresponds to several bytes, an autoregressive byte-level model such as BLT requires multiple decoder forward passes to generate the same amount of text represented by a single subword token. This paper targets that bottleneck. Our goal is to enable byte-level parallel generation while preserving the main benefits of BLT: operating directly on bytes, using dynamic patching, and concentrating computation in latent token representations.

We first draw inspiration from diffusion language models (dLMs), which improve decoding efficiency by generating multiple tokens in parallel within a single forward pass~\citep{sahoo-etal-2024-simple, lou-etal-2024-sedd, wu-etal-2025-fast-dllm, nie-etal-2025-llada, arriola2025block}, reducing memory bandwidth per generated byte.
However, existing text diffusion methods are not directly designed for byte-level architectures whose latent tokens are constructed dynamically from variable-length patches. This creates a key challenge: the model must generate future bytes in parallel while remaining
compatible with BLT’s dynamic, hierarchical architecture.

We introduce \textbf{BLT Diffusion (BLT-D)} (\Cref{fig:inference}), a new byte-level model that combines BLT’s hierarchical latent tokenization with block-wise discrete diffusion. BLT-D retains BLT’s local encoder and global model structure, but modifies training and decoding so that the local decoder can generate a fixed-size block of future bytes in parallel. During training, BLT-D's decoder receives both a clean byte sequence and a corrupted sequence of fixed-length byte blocks. These blocks are constructed from dynamically segmented patches but can extend beyond individual patch boundaries, allowing the decoder to learn to predict future bytes beyond the average BLT patch size. The decoder is trained with a combined objective: the standard autoregressive next-byte prediction loss on clean bytes, and a masked-byte prediction loss on corrupted byte blocks. At inference time, BLT-D initializes a block of masked byte positions and iteratively unmasks multiple positions per decoder step, conditioning on the most recent latent representation. This reduces the number of required decoder, encoder, and global model evaluations per generated sequence.

\begin{figure}[]
    \centering
    \includegraphics[height=0.35\textheight]{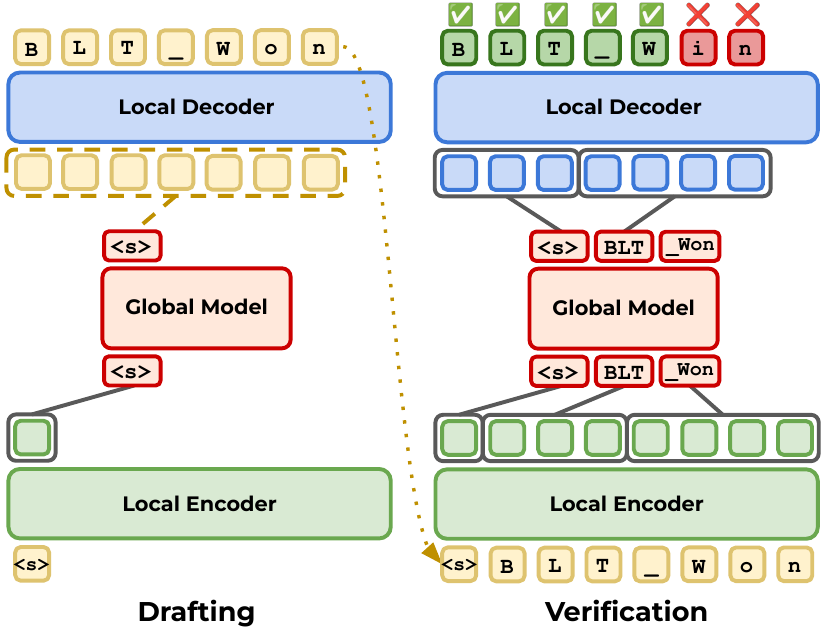}
    \caption{\textbf{Verification procedure shared by BLT-S and BLT-DV.} After a block of bytes is drafted (via self-speculation in BLT-S or diffusion in BLT-DV), the full model re-encodes the candidate sequence and produces next-byte predictions using causal attention. Drafted bytes are accepted up to the first mismatch, which is replaced with the model's prediction. Under greedy decoding, this guarantees that verified outputs are identical to standard autoregressive decoding.
    }
    \label{fig:verification}
\end{figure}

BLT-D offers the largest speedups, but diffusion-based generation introduces a quality–efficiency trade-off. Larger diffusion blocks can reduce inference cost dramatically, because more bytes are generated per decoder call, but they also require the model to predict farther into the future without fully autoregressive conditioning, which can degrade generation quality. To address this, we introduce two additional inference extensions inspired by speculative decoding~\citep{leviathan-etal-2023-speculative, zhang-etal-2024-draft, medusa}. Unlike prior speculative decoding methods that typically use a separate draft model or additional speculative layers, our methods exploit the existing hierarchical structure of BLT and BLT-D (\Cref{fig:verification}). 

The first extension is \textbf{BLT Self-speculation (BLT-S)}. In standard BLT generation, the local decoder stops generating whenever the entropy-based patcher determines that a new patch should begin. BLT-S instead allows the lightweight decoder to autoregressively draft several bytes beyond the usual patch boundary. The full BLT model then verifies this draft using a normal forward pass. If the drafted bytes match the model’s verified predictions, they are accepted; otherwise, generation rolls back to the first mismatch and continues from the verified byte. BLT-S therefore reduces the number of expensive encoder/global calls while preserving the output of standard autoregressive BLT decoding. Unlike conventional speculative decoding, BLT-S does not require a separate draft model: the existing local decoder acts as the drafting mechanism.

The second extension is \textbf{BLT Diffusion+Verification (BLT-DV)}. BLT-D is trained not only with a diffusion objective but also with a standard next-byte prediction objective, so the same model can be run autoregressively with causal decoder masks. BLT-DV uses this fact to combine fast diffusion drafting with autoregressive verification. The diffusion decoder first proposes a block of bytes, and the model then verifies the proposed block using next-byte predictions. This improves generation quality relative to diffusion-only BLT-D while retaining much of the speedup from block-level drafting. BLT-DV therefore occupies a middle point in the trade-off: it is slower than pure BLT-D but typically stronger in task performance.

\paragraph{Contributions}
This paper makes three main contributions:
\begin{enumerate}
    \item We introduce \textbf{BLT-D}, a byte-level language model that makes block-wise discrete diffusion compatible with BLT's dynamic patching and hierarchical latent representations, enabling parallel byte generation without fixed subword tokenization.
    \item We propose two verification-based inference extensions: \textbf{BLT-S}, which accelerates standard BLT using its own decoder as a draft mechanism, and \textbf{BLT-DV}, which improves BLT-D generation quality by verifying diffusion drafts with autoregressive next-byte predictions.
    \item We empirically characterize the speed--quality trade-offs of these methods at 1B and 3B parameter scales across translation and code generation tasks. We provide additional likelihood-based evaluations and generation-diversity analyses.
\end{enumerate}

Across our experiments, BLT-D is our fastest model and inference method, achieving over 50\% lower estimated memory-bandwidth cost compared to BLT on translation and code generation tasks. With larger diffusion block sizes, BLT-D may achieve up to 92\% reduction, with some degradation in task performance. BLT-DV recovers some of this performance while still achieving up to 81\% reduction compared to BLT, and BLT-S achieves up to 77\% reduction with no loss in task performance. Overall, each of these methods has its own unique advantages and helps to further close the inference efficiency gap between byte-level and subword-level models.

\section{Background and Related Work}
\label{sec:background}

In this section, we provide background on BLT and diffusion language models. We further discuss speculative decoding in \Cref{sec:extensions}, where we introduce our extensions.

\subsection{Byte Latent Transformer}
\label{sec:background-blt}

BLT is a byte-level architecture that operates directly on raw byte sequences while matching the performance of subword tokenization-based language models at scale. BLT dynamically groups bytes into variable-length \emph{patches}, which serve as the primary units of computation.
Patches are constructed using an entropy-based segmentation strategy driven by next-byte uncertainty estimated by a small auxiliary byte-level language model. Given a byte input sequence $x = [x_1; x_2; \dots ; x_N] \in \mathcal{V}^N$ of length $N$, where $\mathcal{V}$ is a small byte vocabulary, the sequence is split into $M \approx \frac{N}{4}$ variable-length patches $[p_1; p_2; \dots ; p_M]$. High-entropy regions are segmented into shorter patches, while more predictable spans are grouped into longer patches, thus controlling how frequently the resource-heavy global model is invoked.

\subsubsection{Architecture overview}
BLT's architecture creates latent token representations that mix byte- and patch-level information. It consists of three components: a local encoder $\mathcal{E}$, a global transformer $\mathcal{G}$, and a local decoder $\mathcal{D}$. The local encoder embeds the length-$N$ byte input $x$ to create initial byte representations $\mathbf{X} = [\mathbf{x}_1; \mathbf{x}_2; \dots; \mathbf{x}_{N}] \in \mathbb{R}^{N \times d_{\text{local}}}$, where $d_{\text{local}}$ is the hidden dimensionality of the local encoder and decoder modules and where $\mathbf{x}_i$ is the embedding of byte $x_i$. The encoder then processes $\mathbf{X}$ into $M$ latent token representations $\mathbf{T} =[\mathbf{t}_1; \mathbf{t}_2; \dots ; \mathbf{t}_{M}] \in \mathbb{R}^{M \times d_{\text{global}}}$, where $d_{\text{global}}$ is the hidden dimensionality of the global model. The global Transformer then maps $\mathbf{T}$ to output latent token representations $\mathbf{O} =[\mathbf{o}_1; \mathbf{o}_2; \dots ; \mathbf{o}_{M}] \in \mathbb{R}^{M \times d_{\text{global}}}$. Since our method modifies the decoder, we omit further details of $\mathcal{E}$ and $\mathcal{G}$ and refer the reader to \citealt{pagnoni-etal-2025-byte}.

\subsubsection{Local decoder}
The local decoder $\mathcal{D}$ autoregressively decodes the final latent token representations $\mathbf{o}$ into a sequence of output bytes $y = [y_1; y_2; \dots ; y_N] \in \mathcal{V}^N$ using $L_{\mathcal{D}}$ lightweight Transformer layers. At each layer, byte-level hidden states are updated via cross-attention to latent token representations before applying a standard Transformer layer.
Let $\mathbf{D}_l = [\mathbf{d}_{l,1}; \mathbf{d}_{l,2}; \dots; \mathbf{d}_{l,N}] \in \mathbb{R}^{N \times d_{\text{local}}}$ denote the byte hidden states of a length-$N$ byte sequence output by layer $l$ of the decoder, with $\mathbf{D}_0 \in \mathbb{R}^{N \times d_{\text{local}}}$ being the initial representations from an embedding lookup for $y$.
For each decoder layer $l \in \{1, \dots, L_{\mathcal{D}}\}$, the cross-attention from byte hidden states to latent token representations is computed as
\begin{equation}
\mathbf{B}_l = \mathbf{D}_{l-1} + \mathbf{W}_o \left( \mathrm{softmax}\!\left( \frac{\mathbf{QK}^{\top}}{\sqrt{d_k}} \right) \mathbf{V} \right),
\end{equation}
where $\mathbf{Q}_i = \mathbf{W}_q(\mathbf{d}_{l-1,i})$, $\mathbf{K}_j = \mathbf{W}_k(D_C(\mathbf{o}_j))$, and $\mathbf{V}_j = \mathbf{W}_v(D_C(\mathbf{o}_j))$. Here, $d_k$ is the dimensionality of the key vectors for a single attention head. $\mathbf{W}_q$, $\mathbf{W}_k$, and $\mathbf{W}_v$ are the query, key, and value projection matrices, $D_C(\cdot)$ denotes a linear transformation and splitting function applied to latent token representations, and $\mathbf{W}_o$ is the output projection. The cross-attention does not use positional encodings. The updated byte representations are then produced by
\begin{equation}
\mathbf{D}_l = \operatorname{DecoderTransformerLayer}(\mathbf{B}_l).
\end{equation}
The decoder Transformer layer employs multi-head attention, pre-LayerNorm, and RoPE positional encodings.

\subsection{Diffusion Language Models}
\label{sec:prelim-diffusion}

Diffusion models define generative distributions by progressively corrupting data through a forward noising process and learning a reverse process that iteratively removes noise. Recent work extends this framework to discrete domains such as text by defining stochastic corruption processes over token sequences, enabling training of diffusion language models (dLMs) with diffusion-style objectives and generation over discrete tokens~\citep{austin-etal-2021-structured, campbell-etal-2022-continuous, li-etal-2022-diffusion-lm, gulrajani2023likelihoodbased, lou-etal-2024-sedd}. These models are typically non-autoregressive, employing bidirectional attention over all tokens, or semi-autoregressive, using bidirectional attention within fixed-length blocks while maintaining causal dependencies across blocks~\citep{arriola2025block,gat2025setblockdecodinglanguage}.
Here, we focus on absorbing discrete diffusion with conventions similar to those presented by~\citet{ye2025dream7b} and~\citet{nie-etal-2025-llada}, which is conceptually very similar to masked language models~\citep{devlin-etal-2019-bert}. 

\subsubsection{Absorbing Discrete Diffusion}
We draw a clean text sequence $x^0=[x^0_1; x_2^0 \dots; x^0_N] \in\mathcal{V}^N$ from the data distribution, where $\mathcal{V}$ is the vocabulary and $N$ is the sequence length. We define a discrete diffusion process based on random input masking: given $x^0$, we sample a continuous diffusion timestep (noise level) $t\sim\mathcal{U}(0,1)$ and independently replace each position with a special $\mathtt{[MASK]}$ token with probability $t$, producing a corrupted sequence $x^t$. The forward corruption distribution $q$ is
\begin{equation}
q(x^t_i=\mathtt{[MASK]}\mid x^0_i)=t,\quad
q(x^t_i=x^0_i\mid x^0_i)=1-t,
\label{eq:forward}
\end{equation}
with independence across positions. Prior work has shown that this masking process can be interpreted as the marginal of a discrete diffusion model with an absorbing state, where $\mathtt{[MASK]}$ is absorbing and $t$ controls the diffusion time.

We parameterize a denoising model $p_\theta(x^0_i\mid x^t, t)$ that predicts the original token values at masked positions, conditioned on the partially observed sequence and the noise level. Training minimizes the weighted denoising objective
\begin{align}
\mathcal{L}(\theta)
    &= -\mathbb{E}_{x^0,\,t,\,x^t}
    \Bigg[ \frac{1}{t} \sum_{i=1}^{N} \mathbbm{1}_{\left[x^t_i=\mathtt{[MASK]}\right]} \log p_\theta(x^0_i\mid x^t, t) \Bigg],
    \label{eq:loss}
\end{align}
which has been shown to correspond to a simplified evidence lower bound (ELBO) on the data log-likelihood, or equivalently, an upper bound on the negative log-likelihood~\citep{shi-etal-2024-simplified, gong2025scaling}. Following~\citet{ye2025dream7b} and~\citet{nie-etal-2025-llada}, we do not embed the timestep $t$ into the architecture directly and instead assume that it is implicitly encoded through the input data corruption.

\section{BLT Diffusion}

BLT achieves scalable and efficient byte-level modeling by dynamically allocating compute resources through hierarchical latent tokenization. However, inference speed remains a significant bottleneck, as traditional autoregressive generation proceeds one byte at a time. BLT-D directly addresses this challenge by introducing block diffusion decoding in a way that is fully compatible with BLT’s hierarchical architecture, reducing model calls and therefore memory bandwidth at inference. We adapt the absorbing diffusion framework from ~\Cref{sec:prelim-diffusion} to operate over fixed-size blocks within BLT's decoder.

\subsection{BLT-D Inference}
\label{sec:inference}

BLT-D inference decodes a fully masked block in parallel in much fewer iterations than autoregressively generating a byte at a time (\Cref{fig:inference}).
BLT-D's encoder $\mathcal{E}$ and global model $\mathcal{G}$ operate exactly like BLT, as described in~\Cref{sec:background-blt}. Given a length-$N$ prefix $x = [x_1; \dots; x_N] \in \mathcal{V}^N$, the patcher segments $x$ into $M$ variable-length patches. The encoder $\mathcal{E}$ produces byte embeddings $\mathbf{X} \in \mathbb{R}^{N \times d_{\text{local}}}$ and encodes them into latent token representations $\mathbf{T} = [\mathbf{t}_1; \dots; \mathbf{t}_M] \in \mathbb{R}^{M \times d_{\text{global}}}$. The global model $\mathcal{G}$ outputs contextual latent tokens $\mathbf{O} = [\mathbf{o}_1; \dots; \mathbf{o}_M] \in \mathbb{R}^{M \times d_{\text{global}}}$.

For block diffusion inference, the decoder $\mathcal{D}$ receives as input both the latent token representations $\mathbf{O}$ and a byte sequence $x' = [x_1; \dots; x_N; x_{N+1}; \dots; x_{N+B}] \in \mathcal{V}^{N+B}$, where $[x_{N+1}; \dots; x_{N+B}] = \{\mathtt{[MASK]}\}^B$ form a block of $B$ masked positions. $\mathcal{D}$ iteratively computes forward passes over $x'$ until the entire block of $B$ bytes is unmasked.
See~\Cref{alg:inference} for a more detailed description of the generation procedure.\footnote{The $\mathtt{do\_verify}$ branch is used for BLT-DV, introduced in~\Cref{sec:extensions}; for BLT-D, $\mathtt{do\_verify = False}$.} The subsequent sections detail the inference attention patterns and block unmasking strategies used during generation.

\begin{algorithm}[t]
  \caption{$\mathtt{BLTDGeneration}(x, L, B, \mathtt{do\_verify})$}
  \label{alg:inference}
  \begin{algorithmic}
    \State \textbf{Input:} Initial byte sequence $x = [x_1; \dots; x_N]$; generation length $L$; block size $B$; boolean $\mathtt{do\_verify}$
    \State $l \gets |x|$
    \While{$l < N + L$}
      \State \textbf{Patch Encoding:}
      \State Segment $x$ into $M$ patches via entropy-based patcher
      \State $\mathbf{T} \gets \mathcal{E}(x)$; $\mathbf{O} \gets \mathcal{G}(\mathbf{T})$
      \State \textbf{Block Diffusion Decoding:}
      \State $x_{\text{block}} \gets \{\mathtt{[MASK]}\}^B$
      \State $x' \gets [x_1; \dots; x_l; x_{\text{block}}]$
      \While{$x'$ contains $\mathtt{[MASK]}$}
        \State $y \gets \mathcal{D}(x'; \mathbf{O})$ \Comment{Bidirectional self-attention for block positions}
        \State Select $1 \leq k \leq B$ block positions to unmask \Comment{EB sampling or confidence-based}
        \State Replace selected $\mathtt{[MASK]}$ positions in $x'$ with predictions from $y$
      \EndWhile
      \If{$\mathtt{do\_verify}$}
        \State $x \gets \mathtt{Verify}(x, x', l, B)$
      \Else
        \State $x \gets x'$
      \EndIf
      \State $l \gets |x|$
    \EndWhile
    \State \textbf{Output:} Generated sequence $x$ of length $\geq N + L$
  \end{algorithmic}
\end{algorithm}

\subsubsection{Attention Patterns}

\begin{figure}[htbp]
    \centering
    \includegraphics[width=0.5\columnwidth]{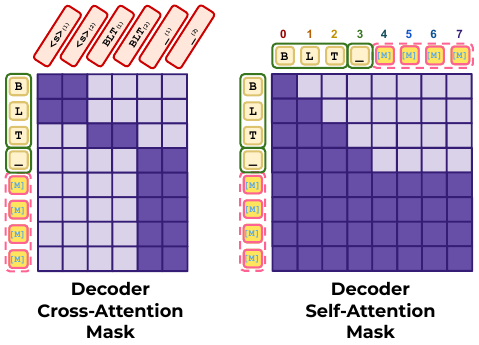}
    \caption{\textbf{BLT-D attention masks during generation with block diffusion.} Before cross-attention, latent tokens are first split into multiple representations via a linear transformation and splitting function (described in detail in~\citealt{pagnoni-etal-2025-byte}). Within the cross-attention, each byte attends to the representations of the previous latent token, except for the last byte of a patch, which may attend to its own latent token. In the self-attention, the clean prefix uses causal attention, and the corrupted/masked portion of the sequence uses bidirectional attention.}
    \label{fig:inference_attn_masks}
\end{figure}

Let $i \in \{1, \dots, N+B\}$ index positions in $x'$. Let $p(i)$ denote the patch index for position $i$ in $x'$. For the decoder's cross-attention module, for clean positions in the sequence ($i \leq N$), each position attends to the latent token $\mathbf{o}_{p(i)-1}$ corresponding to the previous patch, except for the final byte of each patch, which attends to its own latent token $\mathbf{o}_{p(i)}$ (consistent with BLT). For positions in the masked block ($i > N$), all positions attend to the last latent token $\mathbf{o}_{M}$. For $\mathcal{D}$'s self-attention, the attention mask $A \in \{0,1\}^{(N+B) \times (N+B)}$ is defined as follows. For prefix positions ($i \leq N$), $\mathcal{D}$'s self-attention is causal: $A_{ij} = 1$ if $j \leq i$. For block positions ($i > N$), self-attention is fully bidirectional: $A_{ij} = 1$ for all $j \leq N + B$. We provide a visualization of these inference attention masks in~\Cref{fig:inference_attn_masks}.

\subsubsection{Block Unmasking Strategy}
\label{sec:block-unmasking-strategy}

The choice of which bytes to unmask at each decoder forward pass affects both the generation quality and the degree of parallelism. We consider two unmasking strategies that differ in how they select masked positions for decoding.

\paragraph{Confidence-based Unmasking}
The first strategy is confidence-based unmasking~\citep{ghazvininejad-etal-2019-mask}. At each decoder step, the model predicts a distribution over the byte vocabulary for each masked position, and we measure confidence using the maximum predicted probability. All masked positions whose confidence exceeds a threshold $\alpha$ are decoded in parallel, while lower-confidence positions remain masked for subsequent steps. This approach prioritizes high-certainty predictions.
If no position satisfies the threshold, the highest-confidence position is unmasked to ensure progress.

\paragraph{Entropy-bounded Sampling}
The second strategy is entropy-bounded (EB) sampling~\citep{ben-hamu2025accelerated, gat2025setblockdecodinglanguage}. At each decoder step, we compute the entropy of the predicted distribution for each masked token and sort masked positions in ascending order of entropy. Since mutual information among masked tokens is intractable to compute directly, we use an upper bound based on marginal entropies and select the largest subset of positions whose cumulative entropy does not exceed a threshold $\gamma$. The selected tokens are decoded in parallel, while the remaining tokens remain masked. This unmasking strategy may be combined with top-$p$ sampling to obtain diverse generations from the model. Like confidence-based unmasking, if no position satisfies the threshold, the lowest-entropy position is unmasked to ensure progress.

\subsubsection{Speedup}

Compared to standard autoregressive decoding, this approach reduces the number of decoder forward passes: generating a block of size $B$ requires $s$ unmasking steps rather than $B$ sequential steps. Usually, $s < B$, which results in a speedup. Additionally, the encoder and global model are invoked less frequently, as these components are called once per block---typically larger than the average patch---rather than at every new patch. Furthermore, the clean prefix and the first $M-1$ latent tokens from $\mathcal{E}$, $\mathcal{G}$, and $\mathcal{D}$ can be cached, with only the final latent token and drafted block requiring recomputation.

\subsection{BLT-D Training}

BLT-D uses a new training method that enables byte diffusion decoding over latent tokens using specific training data preprocessing, special attention masking in its decoder, and a new loss function. These additions enable BLT-D to predict diffusion blocks that span future bytes far beyond BLT's typical patch size.

\subsubsection{Training Data Preprocessing}

\begin{figure}[]
    \centering
    \includegraphics[width=0.5\columnwidth]{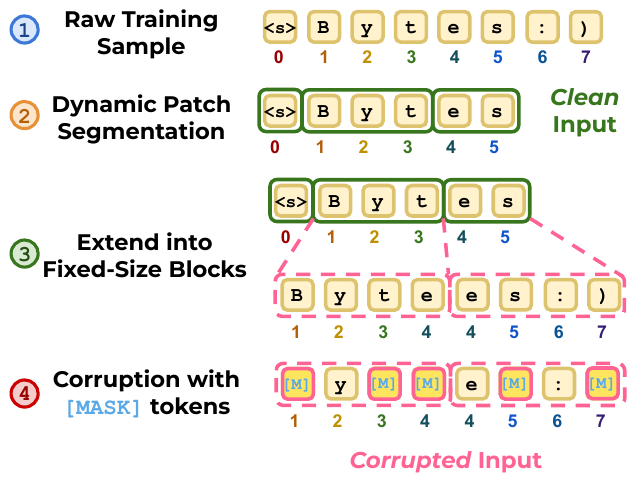}
    \caption{\textbf{BLT-D training data preprocessing.}
    (1) A raw training sample is loaded.
    (2) The entropy patcher segments the input dynamically; for illustration, only the first three patches are shown. This is referred to as the \emph{clean} input.
    (3) All patches except the first are expanded into fixed-size blocks, containing bytes from future patches, with the original positional indices preserved. Allowing predictions beyond the patch enables BLT-D to draft beyond its average patch size during inference.
    (4) The blocks are corrupted with $\mathtt{[MASK]}$s, resulting in the \emph{corrupted} input.}
    \label{fig:train-data-prep}
\end{figure}

To enable block-wise masked prediction, we preprocess each training example as follows. We are given an input byte sequence $x = [x_1; x_2; \dots; x_{N}] \in \mathcal{V}^N$ (where $\mathcal{V}$ is a small byte vocabulary), segmented into $M$ variable-length \emph{patches} with patch $p_i$ starting at index $s_i$.\footnote{Patch $p_1$ is one byte, and is excluded from block construction.}
We construct blocks of bytes and noise these blocks with diffusion, as described in the next paragraphs. For reference, \Cref{fig:train-data-prep} visualizes this data preprocessing for a short example with block size $B=4$.

\paragraph{Block Construction}
From $x$, we construct a corresponding sequence $x_\text{block}$ consisting of $M-1$ fixed-length \emph{blocks} of size $B$. For each patch $p_i$ (excluding the first), we define block $b_{i-1}$ as the $B$ consecutive bytes starting at index $s_i$; that is, for $i \in \{2, \ldots, M\}$, $b_{i-1} = [x_{s_i}; x_{s_i+1}; \dots; x_{s_i+B-1}] \in \mathcal{V}^B$. Since we typically configure $B$ to be greater than the average patch size, these blocks often extend into positions beyond their corresponding patch. This enables BLT-D to predict bytes beyond its average patch size during inference.
If a block extends beyond the end of the sequence ($s_i + B - 1 > N$), we pad it to length $B$ with a special token (e.g. $\mathtt{[PAD]}$).
All blocks are concatenated to form the sequence $x_\text{block} = [b_1; b_2; \dots; b_{M-1}] \in \mathcal{V}^{B \cdot (M-1)}$.
For each block $b_{i-1}$, we record the original byte positional indices $[s_{i}; s_{i}+1; \ldots; s_{i}+B-1]$. These are concatenated for RoPE positional encodings in the decoder during training, ensuring each byte retains representations based on its original position.

\paragraph{Diffusion Process and Masking}
To simulate the diffusion process, we sample a continuous timestep $t \sim \mathcal{U}(0,1)$ and independently replace each byte of $x_\text{block}$ with a $\mathtt{[MASK]}$ token with probability $t$, to produce $x_\text{block}^t \in \{\mathcal{V} \cup \mathtt{[MASK]}\}^{B \cdot (M-1)}$. This produces a partially masked input for the model to reconstruct. We refer to the original sequence $x$ as the \emph{clean} sequence, and $x_\text{block}^t$ as the \emph{corrupted} sequence.

\subsubsection{Decoder Architecture and Attention Patterns}

\begin{figure*}[]
    \centering
    \includegraphics[width=\textwidth]{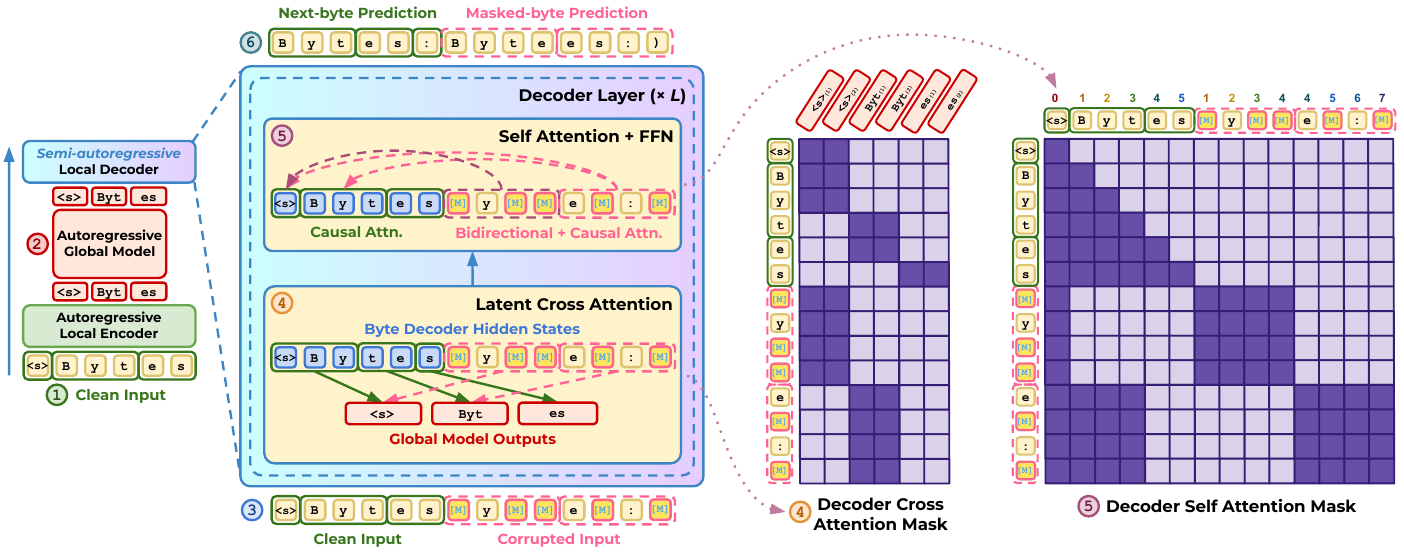}
    \caption{
    \textbf{BLT-D training forward pass.}
    (1-2) The encoder and global model process the clean input.
    (3) Clean and corrupted inputs are concatenated and passed to the decoder.
    (4) Byte hidden states cross-attend to their corresponding latent representations from the global model.
    (5) For the clean portion, self-attention is causal; for the corrupted portion, self-attention is bidirectional within each block, and causal towards previous clean patches.
    (6) Next-byte prediction loss is computed for the clean sequence, and masked byte prediction/diffusion loss is computed for corrupted sequence. 
    }
    \label{fig:training-diagram}
\end{figure*}

The primary architectural innovation in BLT-D lies in the local decoder $\mathcal{D}$, which enables block diffusion decoding. A detailed visualization of the architecture during a training forward pass is shown in \Cref{fig:training-diagram}. BLT-D initializes the decoder input $\mathbf{D}_0$ from embeddings of the concatenated clean and corrupted sequences: $\mathbf{D}_0 = \mathrm{Embed}([x; x_\text{block}^t])$.

For each byte in $\mathbf{D}_0$, cross-attention is applied to the corresponding output latent token in $\mathbf{O}$. Clean sequence positions associated with patch $p_i$ cross-attend to the previous latent token $\mathbf{o}_{i-1}$, except final bytes, which attend to their own latent token $\mathbf{o}_{i}$, consistent with BLT. Corrupted sequence positions associated with patch $p_i$ cross-attend to the previous latent token $\mathbf{o}_{i-1}$. This pattern maintains the alignment between patches and blocks throughout the sequence.
Self-attention in $\mathcal{D}$ uses a causal mask for the clean sequence and bidirectional attention within each block of the corrupted sequence. Each byte within a given block in the corrupted sequence also attends causally to all previous clean bytes. RoPE positional encoding uses the original positional indices as we defined previously.

\subsubsection{Loss Function}

We use a loss function that combines next-byte prediction on the clean sequence with masked reconstruction on the corrupted sequence.
First, recall the clean sequence $x = [x_1; \dots; x_N] \in \mathcal{V}^N$, segmented into $M$ patches with starting indices $s_i$. We compute an autoregressive next-byte prediction loss:
\begin{align}
\mathcal{L}_{\text{clean}}(\theta)
    = - \sum_{i=1}^{N} \log p_\theta(x_i \mid x_{<i})
\end{align}
Here, $p_\theta(x_i \mid x_{<i})$ denotes the model's predicted probability of byte $x_i$ given the prefix $x_{<i}$.
Next, recall the corrupted sequence $x_{\text{block}}^t = [b_1^t; \dots; b_{M-1}^t]$. Each corrupted block $b_{i-1}^t = [x_{s_i}^t; x_{s_i+1}^t; \dots; x_{s_i+B-1}^t] \in \{\mathcal{V} \cup \mathtt{[MASK]}\}^B$ for $i \in \{2, \ldots, M\}$, with each byte masked with probability $t$.
For each corrupted block $b_{i-1}^t$, let $b_{i-1,k}^t$ denote the $k$-th byte of the block. The masked diffusion loss is:
\begin{align}
\mathcal{L}&_{\text{mask}}(\theta) =
    - \frac{1}{t} \sum_{i=2}^{M} \sum_{k=0}^{B-1}
    \mathbbm{1}_{[b_{i-1,k}^t = \mathtt{[MASK]}]} \log p_\theta(x_{s_i+k} \mid b_{i-1}^t, x_{<s_i}) 
\end{align}
where $\mathbbm{1}_{[b_{i-1,k}^t = \mathtt{[MASK]}]}$ is an indicator function that is $1$ if the $k$-th byte of block $b_{i-1}^t$ is masked, and $0$ otherwise.
The model reconstructs the clean byte $x_{s_i+k}$, conditioned on the partially masked block and the clean prefix preceding the block, consistent with the self-attention masking pattern described above.
The scaling by $1/t$ follows the absorbing discrete diffusion loss discussed previously in \Cref{sec:prelim-diffusion}.

The total training loss is the sum of the clean sequence loss and the masked diffusion loss:
\begin{align}
\mathcal{L}_{\text{total}}(\theta) = \mathcal{L}_{\text{clean}}(\theta) + \mathcal{L}_{\text{mask}}(\theta) \label{eq:total_loss}
\end{align}
This combined objective encourages the model to learn both autoregressive next-byte prediction and robust reconstruction of masked bytes in block-wise corrupted sequences.
\section{Pre-training and Generation Experiments}

In this section, we detail the architectures and hyperparameters of each BLT and BLT-D model we train, as well as the pre-training dataset and optimization settings. We evaluate our models on four generation tasks and discuss the efficiency metrics and results.

\subsection{Models, Pre-training Data, and Optimization}

We pre-train four model types: one BLT and three BLT-D variants with block sizes of 4, 8, and 16, referred to as BLT-D-4, BLT-D-8, and BLT-D-16, respectively. For each model type, we train both 1B- and 3B-parameter versions. Our 1B BLT and BLT-D models consist of a global model with 1.28 billion parameters, a local encoder with 19 million parameters, and a local decoder with 160 million parameters. Our 3B BLT and BLT-D models include a global model with 2.82 billion parameters, a local encoder with 26 million parameters, and a local decoder with 160 million parameters. All models employ entropy patching, using an average patch size of 4 bytes and a maximum patch size of 8 bytes.
To ensure comparability, all models are trained on the BLT-1T dataset from~\citealt{pagnoni-etal-2025-byte}, which consists of 1 trillion tokens collected from various public sources and includes a subset of the pre-training data released by Datacomp-LM~\citep{li2024datacomplm}. For additional details on model implementation, hyperparameters, and pre-training optimization settings, see~\Cref{app:arch-and-opt-details}.

\subsection{Generation Tasks, Settings, and Metrics}
\label{sec:generation_tasks}

We evaluate our BLT and BLT-D models on four generation tasks: two translation tasks and two coding tasks. For translation, we evaluate French-to-English and German-to-English (4-shot) using the FLORES-101 benchmark~\citep{goyal-etal-2022-flores}, with performance measured by SentencePiece BLEU. For coding, we assess models on HumanEval (0-shot)~\citep{chen2021evaluatinglargelanguagemodels} and MBPP (3-shot)~\citep{austin2021programsynthesislargelanguage}, reporting $\mathtt{pass@1}$ scores. All task-evaluation inference uses greedy decoding. For BLT-D models, we experiment with both confidence-based unmasking and entropy-bounded sampling as diffusion unmasking strategies, conducting hyperparameter sweeps for each.

Efficiency is evaluated using three metrics: (1) the average number of decoder network function evaluations (NFEs, or forward passes) per output sequence; (2) the average number of encoder/global model NFEs per output sequence; and (3) an estimate of the memory bandwidth required for parameter memory loads during evaluation. The total memory bandwidth, measured in gigabytes, is calculated as follows:
\begin{equation}
\frac{b \left[ N_{\text{dec}} \cdot P_{\text{dec}} + N_{\text{enc}} \cdot (P_{\text{enc}} + P_{\text{glob}}) \right]}{10^9}
\end{equation}
Here, $N_{\text{dec}}$ and $N_{\text{enc}}$ represent the average number of function evaluations for the decoder and encoder/global model, respectively. $P_{\text{dec}}$, $P_{\text{enc}}$, and $P_{\text{glob}}$ denote the number of parameters in the decoder, encoder, and global model. The variable $b$ specifies the number of bytes required to represent each parameter; in our calculations, we set $b = 2$ to reflect 16-bit precision. This formulation assumes that evaluations are performed with a small KV cache and batch size, so the memory bandwidth is dominated by loading model weights. Small batch sizes are common in local serving and latency-oriented applications, where execution speed is prioritized over batching efficiency. BLT-D supports KV caching, and therefore benefits from any techniques that reduce KV-cache memory footprint. Alternatively, memory bandwidth may be interpreted as a weighted function of NFEs for each model component.

\subsection{Generation Task Results}
\label{sec:generation_task_results}

\begin{figure*}[t]
    \centering
    \includegraphics[width=\textwidth]{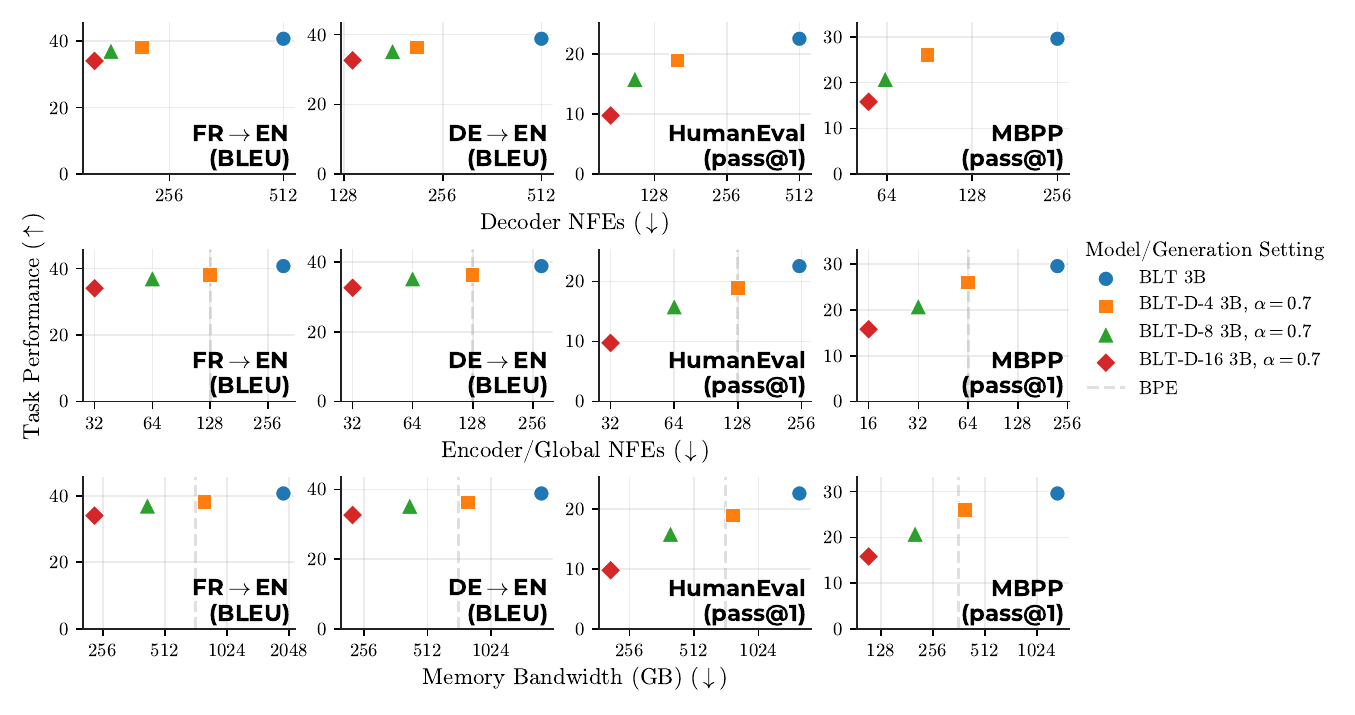}
    \caption{
        Generation task results of 3B-parameter variants of BLT, BLT-D-4, BLT-D-8, and BLT-D-16. Higher is better for task performance; lower is better for NFEs and memory bandwidth. The NFEs and memory bandwidth for a byte-pair encoding (BPE) model matching BLT's global model size are shown as a dashed line. BLT-D models are substantially faster than BLT while maintaining strong task performance, especially for translation. BLT-D-16 offers the most efficiency, with reduced performance on the coding-related tasks.
    }
    \label{fig:main_generation_tasks_3b}
\end{figure*}

We present the performance of our 3B models across a range of generation tasks, as illustrated in \Cref{fig:main_generation_tasks_3b}. For clarity and brevity, this section focuses on representative BLT-D models utilizing confidence-based unmasking as the diffusion generation strategy, with a confidence threshold of $\alpha=0.7$. For comprehensive results for both 1B and 3B model variants, including full inference hyperparameter sweeps using both confidence-based unmasking and EB sampling strategies, please refer to~\Cref{app:all_1b_results} and~\Cref{app:all_3b_results}.

Across all evaluated tasks, BLT-D models consistently outperform BLT in terms of efficiency. Specifically, BLT-D variants achieve substantial reductions in both decoder NFEs and encoder/global NFEs, resulting in large memory bandwidth decreases. For example, BLT-D-4 nearly matches BLT’s task scores while requiring less than half the NFEs and memory bandwidth. Both BLT-D-4 and BLT-D-8 demonstrate strong task performance with great gains in efficiency, especially on the translation tasks.

Increasing the block size in BLT-D models (e.g., BLT-D-16) leads to even greater reductions in NFEs, highlighting the scalability of this approach. BLT-D-16 achieves an 87–92\% reduction in memory bandwidth compared to BLT, making it the fastest model in our evaluations. However, while BLT-D-16 remains competitive on translation tasks, its enhanced efficiency comes at the expense of lower performance on coding-related tasks. This suggests a trade-off between speed and generation quality as block size increases. These results highlight the versatility of BLT-D models, enabling fast generation while allowing flexibility to adjust the block size to suit specific application needs.

\section{Extensions: BLT-S and BLT-DV}
\label{sec:extensions}

Based on our observations from the previous section, BLT achieves strong task performance but suffers from slow generation, while BLT-D greatly improves efficiency but can lose quality at larger block sizes. To improve both models, we draw inspiration from speculative decoding, which accelerates autoregressive generation by separating decoding into a fast \emph{drafting} stage and a slower \emph{verification} stage~\citep{leviathan-etal-2023-speculative}.
In standard speculative decoding, a lightweight \emph{draft model} proposes multiple future tokens, and the large \emph{target model} verifies those proposals in parallel, accepting a prefix of the draft while preserving the target model's output distribution. Subsequent work has reduced the need for a separate draft model by using self-speculation or additional speculative heads~\citep{zhang-etal-2024-draft, medusa}.

Our setting is different: BLT and BLT-D already decompose generation into lightweight byte-level decoding and more expensive encoder/global-model computation. We therefore use the existing model components themselves as drafters: BLT-S drafts with BLT's local decoder beyond normal patch boundaries, while BLT-DV drafts with BLT-D's diffusion decoder and verifies with autoregressive next-byte prediction. These inference extensions require no architectural changes or additional training.

\begin{algorithm}[t]
  \caption{$\mathtt{Verify}(x, x', l, r)$}
  \label{alg:verifyblock}
  \begin{algorithmic}
    \State \textbf{Input:} current sequence $x$; candidate sequence $x'$; start index $l$; draft length $r$
    \State Segment $x'$ into $M'$ patches via entropy-based patcher
    \State $\mathbf{T}' \gets \mathcal{E}(x')$; $\mathbf{O}' \gets \mathcal{G}(\mathbf{T}')$ ; $y \gets \mathcal{D}(x'; \mathbf{O}')$ \Comment{$y_j$ denotes the greedy next-byte prediction after position $j$} 
    \State $i \gets l + 1$
    \While{$i \leq l + r$}
      \If{$x'_i \neq y_{i-1}$}
        \State $x_i \gets y_{i-1}$ \Comment{Reject drafted byte; replace first mismatch}
        \State \textbf{break}
      \Else
        \State $x_i \gets x'_i$ \Comment{Accept drafted byte}
      \EndIf
      \State $i \gets i + 1$
    \EndWhile
    \If{$i = l + r + 1$}
      \State $x_i \gets y_{i-1}$ \Comment{No mismatches; use free byte from next-byte prediction}
    \EndIf
    \State \textbf{Output:} updated sequence $x$
  \end{algorithmic}
\end{algorithm}

\subsection{BLT Self-speculation}
\label{sec:BLT-S-extension}

We introduce a new approach to enhance BLT’s inference efficiency by enabling its decoder to speculate beyond where it would normally segment patches. In standard BLT inference, the entropy-based patcher halts generation whenever a high-entropy byte is produced, prompting a new invocation of the encoder and compute-intensive global model. This patching typically occurs every four bytes.
Instead of immediately patching at each high-entropy byte, we propose a self-speculative decoding strategy, which we call \textbf{BLT-S (BLT Self-speculation)}. Here, the decoder always autoregressively generates up to a fixed window size $k$ regardless of entropy spikes, conditioning on the last available latent token. After producing a draft of $k$ bytes, the patcher segments the sequence and computes a full forward pass through $\mathcal{E}$, $\mathcal{G}$, and $\mathcal{D}$ to obtain new predictions. The model compares the drafted text to these predictions: if all bytes match, the draft is committed; if not, only the bytes up to the first mismatch are accepted. This iterative process advances by at least one verified byte per step and continues until the target sequence length is reached. 

In our setup, verification requires an exact byte-wise match between drafted bytes and the model-verified bytes, and we only evaluate with greedy decoding. This procedure is inspired by speculative decoding but differs in that we validate the bytes themselves rather than their probability distributions; this makes our acceptance criteria stricter than standard speculative decoding. However, our setup is fully compatible with rejection sampling with different temperatures, but we leave explorations of these settings to future work. See~\Cref{alg:verifyblock} for a detailed verification procedure.

Our method fundamentally differs from previous speculative decoding techniques, which typically employ a separate small model or additional layers for self-verification~\citep{leviathan-etal-2023-speculative, zhang-etal-2024-draft, medusa}. In contrast, BLT-S leverages BLT’s existing lightweight decoder ($\mathcal{D}$) for drafting, without introducing auxiliary models or new architectural overhead. 
By allowing the decoder to generate longer speculative windows, BLT-S increases the number of decoder NFEs, but reduces encoder and global model NFEs, leading to improved inference efficiency overall.

\subsection{BLT Diffusion+Verification}
\label{sec:BLT-DV-extension}

Recall that the total training loss in \Cref{eq:total_loss} includes $\mathcal{L}_{\text{clean}}$, the standard autoregressive loss.
Since BLT-D is trained with a next-byte prediction objective, it can be run autoregressively using the same causal decoder masks as BLT. At inference time, the only adjustment needed is to apply the same decoder self-attention and cross-attention masks used in BLT.
This design enables a new generation paradigm for BLT-D, where diffusion acts as the drafting mechanism, while autoregressive next-byte prediction serves as a verification step. We refer to the inference procedure that employs diffusion and verification as \textbf{BLT-DV (BLT Diffusion+Verification)}. After generating a block of bytes via diffusion, BLT-DV performs a full forward pass through $\mathcal{E}$, $\mathcal{G}$, and $\mathcal{D}$ with a causal mask to produce next-byte predictions. The model then verifies the block diffusion draft with the next-byte predictions using the same procedure as in~\Cref{alg:verifyblock}.

Importantly, the same model parameters are used for both drafting and verification. The choice of block size $B$ and unmasking strategy determines the balance between generation speed and verification acceptance rate. Empirically, we found that combining one-step diffusion with verification yields the fastest inference. While one-step diffusion alone typically leads to rapid degradation in generation quality, the verification step effectively prevents this issue.

\begin{figure*}[t]
    \centering
    \includegraphics[width=\textwidth]{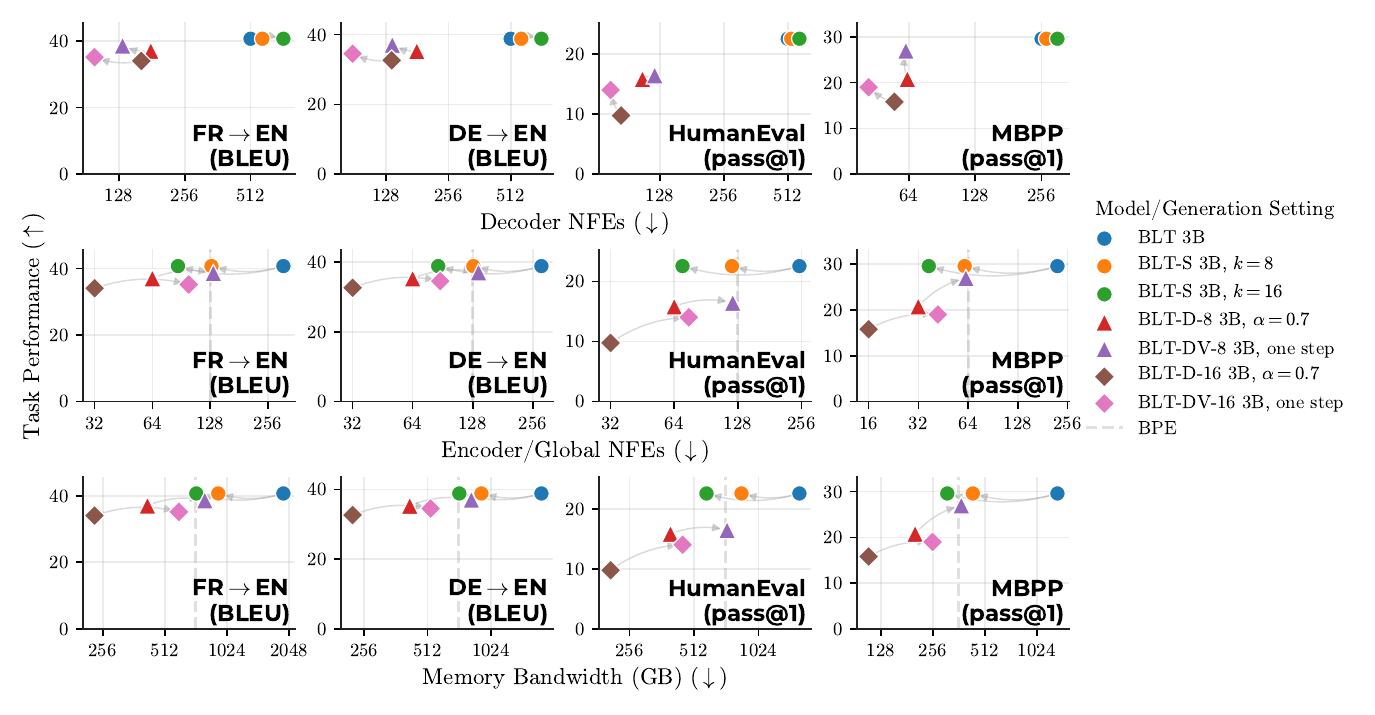}
    \caption{
        Generation task results for 3B-parameter variants of BLT, BLT-S, BLT-D, and BLT-DV. For space, we report results only for $k \in \{8,16\}$ and $B \in \{8,16\}$. Arrows indicate the same model evaluated with different inference methods. Verification (BLT-DV) enhances the task performance of BLT-D models, but increases global NFEs and memory bandwidth. Self-speculation (BLT-S) greatly improves BLT's speed, with no loss in task performance. BLT-D remains the fastest model/inference method overall.
    }
    \label{fig:all_generation_tasks_3b}
\end{figure*}

\subsection{Evaluating Extensions on Generation Tasks}
\label{sec:extension_task_results}

In~\Cref{fig:all_generation_tasks_3b}, we compare 3B-parameter BLT and BLT-D models, along with their respective versions incorporating our new inference extensions. This analysis examines decoder NFEs, encoder/global NFEs, memory bandwidth, and task performance on the same generation tasks described in~\Cref{sec:generation_tasks}. \Cref{app:all_1b_results} and~\Cref{app:all_3b_results} also include additional inference hyperparameter sweeps and generation settings for all 1B and 3B model variants of BLT-S and BLT-DV, along with their verification acceptance rates.
Overall, all BLT-D and BLT-DV outperform BLT and BLT-S in terms of decoder NFEs. Notably, BLT-DV achieves slightly higher task performance than BLT-D without verification; however, this comes at the cost of increased encoder/global NFEs and thus memory bandwidth due to additional verification calls.
BLT-S increases decoder NFEs when compared to BLT, but notably reduces encoder/global NFEs, resulting in improved efficiency and very competitive task performance. Despite these gains, BLT-D-8 and BLT-D-16 (without verification) remain the fastest models, though their task performance is somewhat diminished.

These results suggest several directions for future work. Our experiments used a relatively small decoder; scaling it up could further improve BLT-D/BLT-DV efficiency, since these methods reduce decoder NFEs by design. In contrast, with a lightweight decoder, BLT-S’s extra decoder NFEs impose a smaller overhead, making this approach more appealing. Finally, BLT-DV’s verification may be improved by reweighting the training objective---for example, placing greater emphasis on next-byte prediction (used for verification) relative to diffusion.

\subsection{Likelihood-based Evaluations}

In addition to generation tasks, we further evaluate BLT-D's verification ability on likelihood-based tasks. Since our diffusion models are also trained with a next-byte prediction objective, they inherently possess the ability to compute likelihoods for sequences. By applying a causal mask to the decoder, we can directly obtain these likelihood estimates. Importantly, this serves as a direct proxy for the quality of BLT-DV's verification mechanism, which uses the same masking patterns.
We benchmark the performance of BLT and BLT-D models across five standard datasets: ARC-Easy~\citep{clark2018thinksolvedquestionanswering}, ARC-Challenge~\citep{clark2018thinksolvedquestionanswering}, PIQA~\citep{bisk2019piqareasoningphysicalcommonsense}, HellaSwag~\citep{zellers-etal-2019-hellaswag}, and MMLU~\citep{hendrycks2021measuring} (see \Cref{tab:likelihood_3b}). The results show that BLT-D variants achieve scores approaching those of the BLT baseline, despite the added complexity of balancing next-byte prediction with the diffusion objective. This demonstrates that BLT-D’s autoregressive capabilities remain robust and that the integration of block diffusion does not compromise autoregressive performance on established language understanding and reasoning tasks.
Overall, these findings suggest that BLT-D models can effectively combine block diffusion and next-byte prediction objectives, maintaining strong performance while ensuring high-quality generations.

\begin{table}[]
\centering
\caption{Performance comparison of 3B-parameter BLT and BLT-D models (block sizes: 4, 8, 16) across five benchmarks. While BLT-D variants exhibit a performance hit due to balancing next-byte prediction with the diffusion objective, the diffusion mechanism enables much faster inference for BLT-D.}
\begin{tabular}{lcccc}
\toprule
\textbf{Benchmark} & \textbf{BLT 3B} & \textbf{BLT-D-4 3B} & \textbf{BLT-D-8 3B} & \textbf{BLT-D-16 3B} \\
\midrule
\textbf{ARC-Easy}       & 74.33 & 72.39 & 70.95 & 66.89 \\
\textbf{ARC-Challenge}  & 45.75 & 41.46 & 41.03 & 40.43 \\
\textbf{PIQA}           & 79.38 & 79.60 & 78.02 & 76.93 \\
\textbf{HellaSwag}      & 74.98 & 71.86 & 70.56 & 69.12 \\
\textbf{MMLU}           & 41.15 & 39.07 & 38.29 & 37.08 \\
\bottomrule
\end{tabular}
\label{tab:likelihood_3b}
\end{table}

\section{BLT-D Generation Analysis}
\label{app:generation-analysis}

In this section, we analyze the diversity and efficiency of unconditional generations produced by BLT-D models using entropy-bounded sampling as the unmasking strategy. Because entropy-bounded sampling can be combined with top-$p$ sampling, this setup allows us to sample diverse outputs while varying the amount of parallelism during block diffusion decoding. This analysis focuses on the block generation ability of BLT-D without autoregressive next-byte verification. For each model and sampling configuration, we generate text unconditionally from the start-of-sequence token until reaching a maximum length of 1k bytes.

\begin{figure}[t]
    \centering
    \begin{subfigure}[t]{0.9\textwidth}
    \centering
        \includegraphics[width=\textwidth]{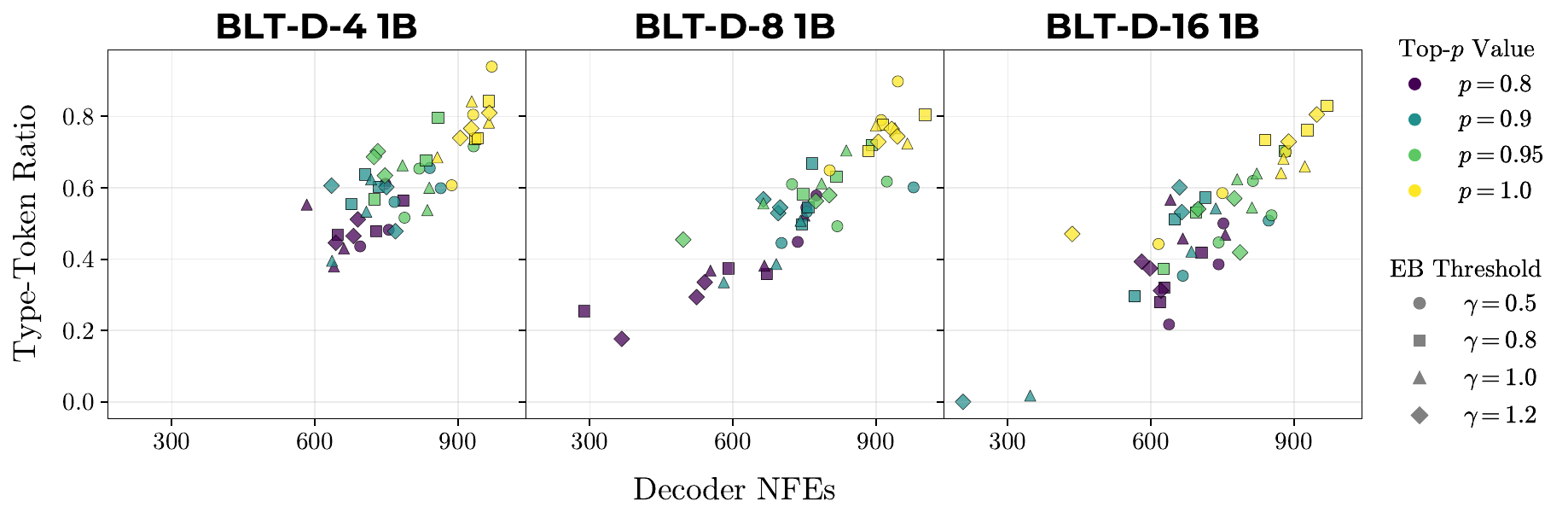}
        \caption{1B models.}
    \end{subfigure}
    \begin{subfigure}[t]{0.9\textwidth}
    \centering
        \includegraphics[width=\textwidth]{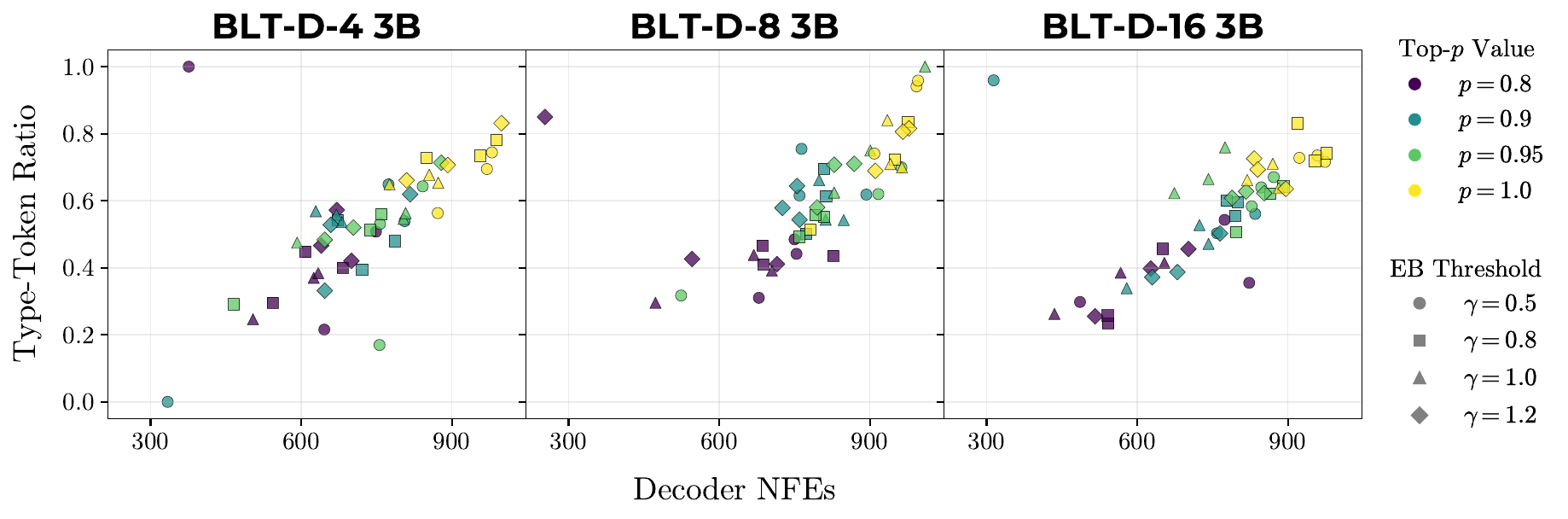}
        \caption{3B models.}
    \end{subfigure}
    \caption{
        Type-token ratio increases with the number of decoder calls when generating text with BLT-D using entropy-bounded sampling with top-$p$ sampling. This indicates that more decoder passes yield greater diversity, while fewer passes correspond to more repetitive, predictable text. Block diffusion decoding enables exploration of this trade-off between generation diversity and computational efficiency.
    }
    \label{fig:ttr_decoder_calls}
\end{figure}

To quantify diversity, we compute the word-level type-token ratio (TTR) of the generated text after whitespace tokenization, and compare it against the number of decoder network function evaluations (NFEs) required under varying entropy-bounded sampling thresholds ($\gamma$) and top-$p$ values. TTR serves as a simple proxy for lexical diversity, with higher values indicating a greater variety of unique words relative to the total word count. The resulting diversity--efficiency trade-off is shown in~\Cref{fig:ttr_decoder_calls}.

Our results show a clear trend: as the number of decoder calls increases, the type-token ratio also increases. This suggests that more decoder forward passes are associated with the generation of more diverse text. Conversely, when the model produces repetitive or highly predictable text, it requires fewer decoder calls, reflecting the lower uncertainty and entropy in those outputs. This relationship highlights a key advantage of block diffusion decoding: it provides a mechanism to explore the trade-off between generation diversity and computational efficiency. 

\section{Conclusion}

In this paper, we introduced \textbf{BLT Diffusion (BLT-D)}, a byte-level language model that combines BLT's hierarchical latent tokenization with a block-wise diffusion objective to accelerate generation. BLT-D's new semi-autoregressive decoder design enables multiple future bytes to be generated in parallel, all while preserving BLT's dynamic patching and latent token representations. We also proposed two speculative-decoding--inspired extensions: \textbf{BLT Self-speculation (BLT-S)}, which uses BLT's own decoder to draft beyond normal patch boundaries before verification, and \textbf{BLT Diffusion+Verification (BLT-DV)}, which verifies diffusion drafts using autoregressive next-byte prediction. Each of these methods substantially reduces total model calls, narrowing the inference-efficiency gap between byte-level and subword-level models.

\paragraph{Limitations and Future Work}
Here, we note our limitations and point out exciting avenues for future work.
The main limitation of our evaluation is that we use network function evaluations (NFEs) and estimated memory bandwidth as proxy metrics for inference efficiency. NFEs are commonly reported in the discrete diffusion literature (see~\citealt{lou-etal-2024-sedd, arriola2025block}) because they isolate algorithmic efficiency from implementation-specific factors such as kernels, hardware utilization, batching strategy, and KV-cache management. Benchmarking BLT, BLT-D, BLT-S, and BLT-DV in a highly optimized inference implementation is therefore an important direction for future work. Other promising directions include experimenting with different patch sizes, tuning the balance between BLT-D's diffusion and next-byte prediction objectives, scaling training further---which may especially benefit diffusion language models~\citep{ni2025diffusionlanguagemodelssuper}---and studying how decoder parameter allocation affects the performance and efficiency of each BLT variant.

\newpage
\bibliographystyle{assets/plainnat}
\bibliography{paper}

\newpage
\beginappendix

\section{Architecture and Optimization Details}
\label{app:arch-and-opt-details}

\subsection{Architecture Implementation Details}

For all the BLT and BLT-D models we train, we maintain the same Transformer implementation details from the original BLT: the feed-forward layers use the SwiGLU activation function \citep{shazeer2020glu}, all self-attention modules use rotary positional embeddings (RoPE, \citealt{su2023roformer}) with $\theta=500000$ \citep{xiong-etal-2024-effective}, and layer normalization is done with RMSNorm \citep{rmsnorm}.

For self-attention in the encoder and global model, where the mask is fixed and follows a standard causal pattern with a fixed window, we use FlashAttention \citep{dao2022flashattention} with a window size of 512. For all cross-attention modules and the decoder's self-attention module, which requires carefully constructed custom masks that depend on the patch structure and vary per example, we use FlexAttention \citep{dong2025flexattention}. FlexAttention streamlines the implementation of attention mechanisms with structured sparsity in PyTorch and allows users to define custom attention masks, all while achieving performance levels on par with specialized, manually optimized attention kernels.

\subsection{Pre-training Optimization and Hyperparameter Settings}

All BLT/BLT-D 1B models are trained for 240{,}000 steps with a batch size of $2^{19}$ tokens per step (approximately 2 million bytes), and our 3B models are trained for 480{,}000 steps with a batch size of $2^{20}$ tokens per step (approximately 4 million bytes).
All models use the AdamW optimizer \citep{adamw} with $\beta_1=0.9$, $\beta_2 = 0.95$, and $\epsilon = 10^{-8}$. All models use a cosine learning rate schedule that linearly warms-up to a peak learning rate of $4 \times 10^{-4}$ and decays to 0. The 1B models warm-up to 2000 steps; the 3B models warm-up to 4000 steps. We apply a weight decay of 0.1, and global gradient clipping at a threshold of 1.0.

\newpage
\section{All 1B Model Results}
\label{app:all_1b_results}

\begin{figure*}[htbp]
    \centering
    \includegraphics[width=\textwidth]{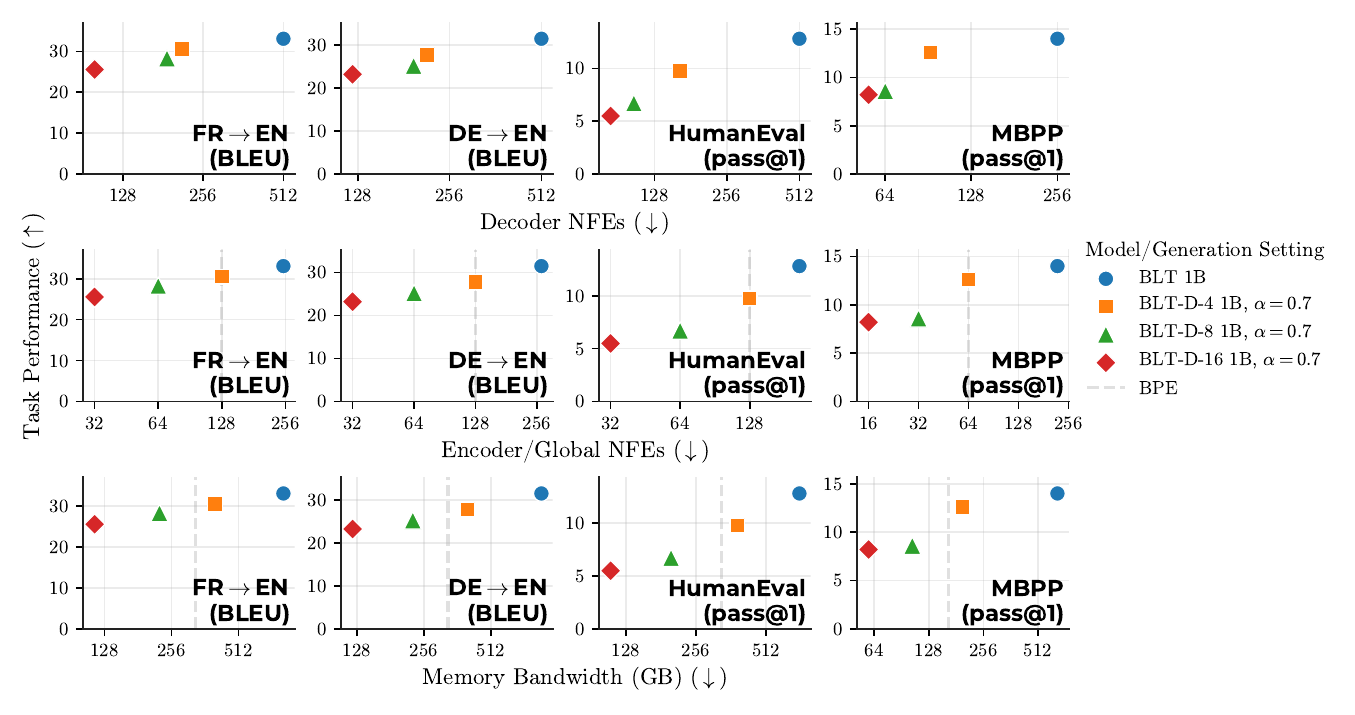}
    \caption{
        Generation task results of 1B-parameter variants of BLT, BLT-D-4, BLT-D-8, and BLT-D-16. Higher is better for task performance; lower is better for NFEs and memory bandwidth. The NFEs and memory bandwidth for a byte-pair encoding (BPE) model matching BLT's global model size are shown as a dashed line.
    }
    \label{fig:main_generation_tasks_1b}
\end{figure*}

\begin{figure*}[htbp]
    \centering
    \includegraphics[width=\textwidth]{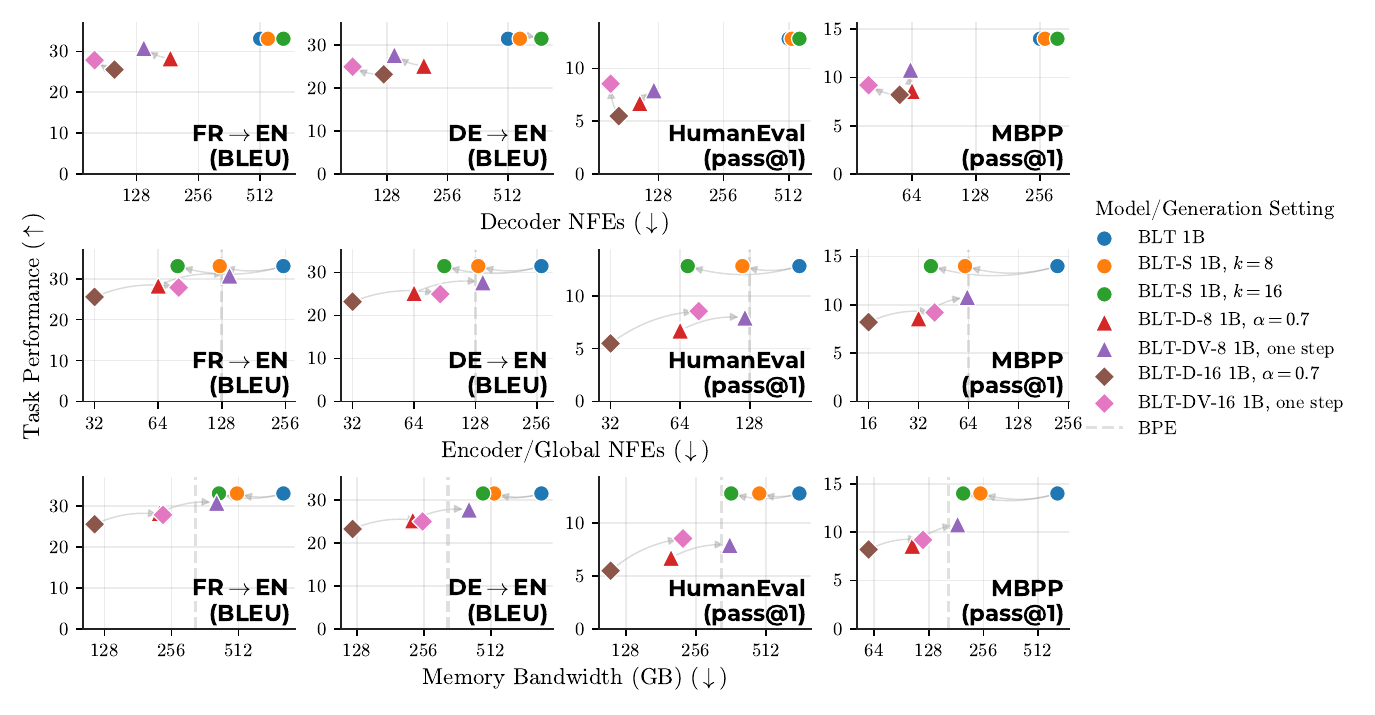}
    \caption{
        Generation task results for 1B-parameter variants of BLT, BLT-S, BLT-D, and BLT-DV. For space, we report results only for $k \in \{8,16\}$ and $B \in \{8,16\}$. Arrows indicate the same model evaluated with different inference methods.
    }
    \label{fig:all_generation_tasks_1b}
\end{figure*}

In this section, we report results for all 1B models. \Cref{fig:main_generation_tasks_1b} and \Cref{fig:all_generation_tasks_1b} present the 1B counterparts of the generation-task results from \Cref{sec:generation_task_results} and \Cref{sec:extension_task_results} for BLT, BLT-D, BLT-S, and BLT-DV. \Cref{tab:likelihood_1b} reports the likelihood-based evaluation results for the 1B models.

We also run a larger sweep over inference hyperparameters for the 1B models on the generation tasks. For BLT-D, we evaluate confidence-based unmasking with thresholds $\alpha \in \{0.5, 0.7\}$, as well as EB sampling with thresholds $\gamma \in \{0.8, 1.0\}$. For BLT-DV, we use more permissive settings that unmask more bytes per step; i.e., we \emph{decrease} $\alpha$ for confidence-based unmasking or \emph{increase} $\gamma$ for EB sampling. Specifically, we test $\alpha = 0.3$, $\gamma \in \{1.5, 2.0\}$, and one-step diffusion that unmasks all byte positions at once. For BLT-S, we use speculation windows $k \in \{4, 8, 16\}$. For BLT-S and BLT-DV, we also report the verification acceptance rate, defined as the fraction of drafted bytes that are accepted after verification. \Cref{tab:fr_en_results_1b}, \Cref{tab:de_en_results_1b}, \Cref{tab:human_eval_1b}, and \Cref{tab:MBPP_results_1b} report results on French-to-English translation, German-to-English translation, HumanEval, and MBPP, respectively.

\begin{table}[htbp]
\centering
\caption{Performance comparison of BLT and BLT-D (block sizes 4, 8, 16) at 1B parameters across five benchmarks.}
\begin{tabular}{lcccc}
\toprule
\textbf{Benchmark} & \textbf{BLT 1B} & \textbf{BLT-D-4 1B} & \textbf{BLT-D-8 1B} & \textbf{BLT-D-16 1B} \\
\midrule
\textbf{ARC-Easy}       & 63.21 & 60.76 & 61.06 & 59.83 \\
\textbf{ARC-Challenge}  & 34.94 & 32.96 & 34.16 & 32.88 \\
\textbf{PIQA}           & 75.46 & 74.48 & 73.56 & 72.36 \\
\textbf{HellaSwag}      & 60.17 & 59.06 & 58.34 & 57.13 \\
\textbf{MMLU}           & 33.90 & 33.60 & 32.28 & 32.09 \\
\bottomrule
\end{tabular}
\label{tab:likelihood_1b}
\end{table}

\begin{table*}[htbp]
\centering
\caption{Full \textbf{French-to-English} translation results for \textbf{1B-parameter} models across various generation settings.}
\resizebox{\textwidth}{!}{%
\begin{tabular}{l l c l c c c c c}
\toprule
\textbf{Model} & \textbf{Generation Setting} & \textbf{BLEU} & \shortstack{\textbf{Diffusion/Speculation} \\ \textbf{Sampling Strategy}} & \shortstack{\textbf{Acceptance} \\ \textbf{Rate (\%)}} & \shortstack{\textbf{Decoder} \\ \textbf{NFEs}} & \shortstack{\textbf{Global} \\ \textbf{NFEs}} & \shortstack{\textbf{Memory} \\ \textbf{Bandwidth} \\ \textbf{(GB)}} & \shortstack{\textbf{Memory} \\ \textbf{Decrease} \\ \textbf{vs. BLT (\%)}} \\
\midrule
\multirow{4}{*}{BLT 1B} & BLT (AR) & 33.08 & --- & --- & 512 & 250 & 814.95 & --- \\
\cmidrule(lr){2-9}
& \multirow{3}{*}{\shortstack[l]{BLT-S \\ (AR+self-speculation)}} & \multirow{3}{*}{33.08} & $k=4$ & 96.77 & 526 & 212 & 719.45 & 11.72 \\
& & & $k=8$ & 91.14 & 558 & 125 & 504.08 & 38.15 \\
& & & $k=16$ & 76.93 & 664 & 79 & 418.26 & 48.68 \\
\cmidrule(lr){1-9}
\multirow{8}{*}{BLT-D-4 1B}
    & \multirow{4}{*}{\shortstack[l]{BLT-D \\ (diffusion only)}} & 30.01 & Confidence-based, $\alpha=0.5$ & --- & 184 & 128 & 392.17 & 51.88 \\
    & & 30.53 & Confidence-based, $\alpha=0.7$ & --- & 213 & 128 & 401.38 & 50.75 \\
    & & 30.59 & EB sampling, $\gamma=0.8$ & --- & 261 & 128 & 416.32 & 48.91 \\
    & & 30.68 & EB sampling, $\gamma=1.0$ & --- & 249 & 128 & 412.68 & 49.36 \\
    \cmidrule(lr){2-9}
    & \multirow{4}{*}{\shortstack[l]{BLT-DV \\ (diffusion+verification)}} & \multirow{4}{*}{32.65} & Confidence-based, $\alpha=0.3$ & 93.40 & 239 & 217 & 642.00 & 21.22 \\
    & & & EB sampling, $\gamma=1.5$ & 94.41 & 299 & 215 & 656.19 & 19.48 \\
    & & & EB sampling, $\gamma=2.0$ & 94.38 & 284 & 215 & 651.69 & 20.03 \\
    & & & one step & 92.55 & 218 & 218 & 639.70 & 21.50 \\
\cmidrule(lr){1-9}
\multirow{8}{*}{BLT-D-8 1B}
    & \multirow{4}{*}{\shortstack[l]{BLT-D \\ (diffusion only)}} & 26.70 & Confidence-based, $\alpha=0.5$ & --- & 147 & 64 & 213.50 & 73.80 \\
    & & 28.32 & Confidence-based, $\alpha=0.7$ & --- & 187 & 64 & 226.00 & 72.27 \\
    & & 28.11 & EB sampling, $\gamma=0.8$ & --- & 259 & 64 & 248.91 & 69.46 \\
    & & 28.14 & EB sampling, $\gamma=1.0$ & --- & 244 & 64 & 243.95 & 70.07 \\
    \cmidrule(lr){2-9}
    & \multirow{4}{*}{\shortstack[l]{BLT-DV \\ (diffusion+verification)}} & \multirow{4}{*}{30.80} & Confidence-based, $\alpha=0.3$ & 83.81 & 176 & 134 & 406.37 & 50.14 \\
    & & & EB sampling, $\gamma=1.5$ & 86.91 & 272 & 130 & 425.34 & 47.81 \\
    & & & EB sampling, $\gamma=2.0$ & 86.69 & 253 & 130 & 420.33 & 48.42 \\
    & & & one step & 80.34 & 139 & 139 & 408.23 & 49.91 \\
\cmidrule(lr){1-9}
\multirow{8}{*}{BLT-D-16 1B}
    & \multirow{4}{*}{\shortstack[l]{BLT-D \\ (diffusion only)}} & 23.68 & Confidence-based, $\alpha=0.5$ & --- & 77 & 32 & 107.92 & 86.76 \\
    & & 25.55 & Confidence-based, $\alpha=0.7$ & --- & 100 & 32 & 115.34 & 85.85 \\
    & & 25.49 & EB sampling, $\gamma=0.8$ & --- & 179 & 32 & 140.24 & 82.79 \\
    & & 25.44 & EB sampling, $\gamma=1.0$ & --- & 167 & 32 & 136.42 & 83.26 \\
    \cmidrule(lr){2-9}
    & \multirow{4}{*}{\shortstack[l]{BLT-DV \\ (diffusion+verification)}} & \multirow{4}{*}{27.84} & Confidence-based, $\alpha=0.3$ & 82.49 & 112 & 74 & 230.57 & 71.71 \\
    & & & EB sampling, $\gamma=1.5$ & 86.53 & 201 & 71 & 248.80 & 69.47 \\
    & & & EB sampling, $\gamma=2.0$ & 86.27 & 184 & 71 & 244.13 & 70.04 \\
    & & & one step & 77.19 & 80 & 80 & 234.33 & 71.25 \\
\bottomrule
\end{tabular}
}
\label{tab:fr_en_results_1b}
\end{table*}

\begin{table*}[htbp]
\centering
\caption{Full \textbf{German-to-English} translation results for \textbf{1B-parameter} models across various generation settings.}
\resizebox{\textwidth}{!}{%
\begin{tabular}{l l c l c c c c c}
\toprule
\textbf{Model} & \textbf{Generation Setting} & \textbf{BLEU} & \shortstack{\textbf{Diffusion/Speculation} \\ \textbf{Sampling Strategy}} & \shortstack{\textbf{Acceptance} \\ \textbf{Rate (\%)}} & \shortstack{\textbf{Decoder} \\ \textbf{NFEs}} & \shortstack{\textbf{Global} \\ \textbf{NFEs}} & \shortstack{\textbf{Memory} \\ \textbf{Bandwidth} \\ \textbf{(GB)}} & \shortstack{\textbf{Memory} \\ \textbf{Decrease} \\ \textbf{vs. BLT (\%)}} \\
\midrule
\multirow{4}{*}{BLT 1B} & BLT (AR) & 31.46 & --- & --- & 512 & 269 & 864.76 & --- \\
\cmidrule(lr){2-9}
& \multirow{3}{*}{\shortstack[l]{BLT-S \\ (AR+self-speculation)}} & \multirow{3}{*}{31.46} & $k=4$ & 95.08 & 534 & 215 & 729.53 & 15.64 \\
& & & $k=8$ & 86.09 & 587 & 132 & 530.01 & 38.71 \\
& & & $k=16$ & 67.51 & 751 & 90 & 472.52 & 45.36 \\
\cmidrule(lr){1-9}
\multirow{8}{*}{BLT-D-4 1B}
    & \multirow{4}{*}{\shortstack[l]{BLT-D \\ (diffusion only)}} & 27.30 & Confidence-based, $\alpha=0.5$ & --- & 189 & 128 & 393.58 & 54.49 \\
    & & 27.73 & Confidence-based, $\alpha=0.7$ & --- & 216 & 128 & 402.31 & 53.48 \\
    & & 27.85 & EB sampling, $\gamma=0.8$ & --- & 259 & 128 & 415.71 & 51.93 \\
    & & 27.99 & EB sampling, $\gamma=1.0$ & --- & 247 & 128 & 411.90 & 52.37 \\
    \cmidrule(lr){2-9}
    & \multirow{4}{*}{\shortstack[l]{BLT-DV \\ (diffusion+verification)}} & \multirow{4}{*}{29.56} & Confidence-based, $\alpha=0.3$ & 93.45 & 238 & 217 & 641.53 & 25.81 \\
    & & & EB sampling, $\gamma=1.5$ & 94.42 & 284 & 215 & 651.55 & 24.66 \\
    & & & EB sampling, $\gamma=2.0$ & 94.33 & 273 & 215 & 648.30 & 25.03 \\
    & & & one step & 93.36 & 217 & 217 & 635.20 & 26.55 \\
\cmidrule(lr){1-9}
\multirow{8}{*}{BLT-D-8 1B}
    & \multirow{4}{*}{\shortstack[l]{BLT-D \\ (diffusion only)}} & 24.20 & Confidence-based, $\alpha=0.5$ & --- & 157 & 64 & 216.67 & 74.94 \\
    & & 25.20 & Confidence-based, $\alpha=0.7$ & --- & 195 & 64 & 228.50 & 73.58 \\
    & & 24.77 & EB sampling, $\gamma=0.8$ & --- & 252 & 64 & 246.55 & 71.49 \\
    & & 24.92 & EB sampling, $\gamma=1.0$ & --- & 238 & 64 & 242.16 & 72.00 \\
    \cmidrule(lr){2-9}
    & \multirow{4}{*}{\shortstack[l]{BLT-DV \\ (diffusion+verification)}} & \multirow{4}{*}{27.71} & Confidence-based, $\alpha=0.3$ & 82.94 & 188 & 135 & 414.05 & 52.12 \\
    & & & EB sampling, $\gamma=1.5$ & 85.35 & 278 & 132 & 433.24 & 49.90 \\
    & & & EB sampling, $\gamma=2.0$ & 85.16 & 260 & 132 & 428.32 & 50.47 \\
    & & & one step & 80.41 & 139 & 139 & 408.96 & 52.71 \\
\cmidrule(lr){1-9}
\multirow{8}{*}{BLT-D-16 1B}
    & \multirow{4}{*}{\shortstack[l]{BLT-D \\ (diffusion only)}} & 21.87 & Confidence-based, $\alpha=0.5$ & --- & 94 & 32 & 113.32 & 86.90 \\
    & & 23.19 & Confidence-based, $\alpha=0.7$ & --- & 123 & 32 & 122.48 & 85.84 \\
    & & 22.66 & EB sampling, $\gamma=0.8$ & --- & 208 & 32 & 149.20 & 82.75 \\
    & & 22.66 & EB sampling, $\gamma=1.0$ & --- & 195 & 32 & 145.24 & 83.20 \\
    \cmidrule(lr){2-9}
    & \multirow{4}{*}{\shortstack[l]{BLT-DV \\ (diffusion+verification)}} & \multirow{4}{*}{24.96} & Confidence-based, $\alpha=0.3$ & 76.51 & 144 & 80 & 256.03 & 70.39 \\
    & & & EB sampling, $\gamma=1.5$ & 80.15 & 245 & 76 & 278.02 & 67.85 \\
    & & & EB sampling, $\gamma=2.0$ & 79.93 & 226 & 77 & 272.54 & 68.48 \\
    & & & one step & 71.75 & 86 & 86 & 252.74 & 70.77 \\
\bottomrule
\end{tabular}
}
\label{tab:de_en_results_1b}
\end{table*}

\begin{table*}[htbp]
\centering
\caption{Full \textbf{HumanEval} task results for \textbf{1B-parameter} models across various generation settings.}
\resizebox{\textwidth}{!}{%
\begin{tabular}{l l c l c c c c c}
\toprule
\textbf{Model} & \textbf{Generation Setting} & \textbf{PASS@1} & \shortstack{\textbf{Diffusion/Speculation} \\ \textbf{Sampling Strategy}} & \shortstack{\textbf{Acceptance} \\ \textbf{Rate (\%)}} & \shortstack{\textbf{Decoder} \\ \textbf{NFEs}} & \shortstack{\textbf{Global} \\ \textbf{NFEs}} & \shortstack{\textbf{Memory} \\ \textbf{Bandwidth} \\ \textbf{(GB)}} & \shortstack{\textbf{Memory} \\ \textbf{Decrease} \\ \textbf{vs. BLT (\%)}} \\
\midrule
\multirow{4}{*}{BLT 1B} & BLT (AR) & 12.80 & --- & --- & 512 & 210 & 711.20 & --- \\
\cmidrule(lr){2-9}
& \multirow{3}{*}{\shortstack[l]{BLT-S \\ (AR+self-speculation)}} & \multirow{3}{*}{12.80} & $k=4$ & 98.77 & 517 & 208 & 707.67 & 0.50 \\
& & & $k=8$ & 96.45 & 529 & 119 & 478.64 & 32.70 \\
& & & $k=16$ & 88.92 & 574 & 69 & 362.77 & 48.99 \\
\cmidrule(lr){1-9}
\multirow{8}{*}{BLT-D-4 1B}
    & \multirow{4}{*}{\shortstack[l]{BLT-D \\ (diffusion only)}} & 8.54 & Confidence-based, $\alpha=0.5$ & --- & 147 & 128 & 380.26 & 46.53 \\
    & & 9.76 & Confidence-based, $\alpha=0.7$ & --- & 163 & 128 & 385.31 & 45.82 \\
    & & 10.37 & EB sampling, $\gamma=0.8$ & --- & 195 & 128 & 395.62 & 44.37 \\
    & & 10.37 & EB sampling, $\gamma=1.0$ & --- & 185 & 128 & 392.50 & 44.81 \\
    \cmidrule(lr){2-9}
    & \multirow{4}{*}{\shortstack[l]{BLT-DV \\ (diffusion+verification)}} & \multirow{4}{*}{9.15} & Confidence-based, $\alpha=0.3$ & 97.76 & 213 & 209 & 613.63 & 13.72 \\
    & & & EB sampling, $\gamma=1.5$ & 98.22 & 239 & 208 & 619.73 & 12.86 \\
    & & & EB sampling, $\gamma=2.0$ & 98.22 & 231 & 208 & 617.23 & 13.21 \\
    & & & one step & 97.69 & 209 & 209 & 612.52 & 13.88 \\
\cmidrule(lr){1-9}
\multirow{8}{*}{BLT-D-8 1B}
    & \multirow{4}{*}{\shortstack[l]{BLT-D \\ (diffusion only)}} & 6.71 & Confidence-based, $\alpha=0.5$ & --- & 87 & 64 & 194.69 & 72.62 \\
    & & 6.71 & Confidence-based, $\alpha=0.7$ & --- & 105 & 64 & 200.35 & 71.83 \\
    & & 7.93 & EB sampling, $\gamma=0.8$ & --- & 155 & 64 & 215.99 & 69.63 \\
    & & 7.93 & EB sampling, $\gamma=1.0$ & --- & 144 & 64 & 212.55 & 70.11 \\
    \cmidrule(lr){2-9}
    & \multirow{4}{*}{\shortstack[l]{BLT-DV \\ (diffusion+verification)}} & \multirow{4}{*}{7.93} & Confidence-based, $\alpha=0.3$ & 93.73 & 131 & 121 & 358.72 & 49.56 \\
    & & & EB sampling, $\gamma=1.5$ & 95.25 & 180 & 119 & 369.54 & 48.04 \\
    & & & EB sampling, $\gamma=2.0$ & 95.15 & 169 & 119 & 366.22 & 48.51 \\
    & & & one step & 93.13 & 122 & 122 & 357.62 & 49.72 \\
\cmidrule(lr){1-9}
\multirow{8}{*}{BLT-D-16 1B}
    & \multirow{4}{*}{\shortstack[l]{BLT-D \\ (diffusion only)}} & 3.66 & Confidence-based, $\alpha=0.5$ & --- & 61 & 32 & 102.82 & 85.54 \\
    & & 5.49 & Confidence-based, $\alpha=0.7$ & --- & 84 & 32 & 110.27 & 84.50 \\
    & & 6.10 & EB sampling, $\gamma=0.8$ & --- & 148 & 32 & 130.41 & 81.66 \\
    & & 5.49 & EB sampling, $\gamma=1.0$ & --- & 137 & 32 & 126.72 & 82.18 \\
    \cmidrule(lr){2-9}
    & \multirow{4}{*}{\shortstack[l]{BLT-DV \\ (diffusion+verification)}} & \multirow{4}{*}{8.54} & Confidence-based, $\alpha=0.3$ & 81.72 & 97 & 74 & 225.05 & 68.36 \\
    & & & EB sampling, $\gamma=1.5$ & 86.99 & 180 & 70 & 240.02 & 66.25 \\
    & & & EB sampling, $\gamma=2.0$ & 86.75 & 164 & 70 & 235.13 & 66.94 \\
    & & & one step & 78.92 & 77 & 77 & 225.34 & 68.32 \\
\bottomrule
\end{tabular}
}
\label{tab:human_eval_1b}
\end{table*}

\begin{table*}[htbp]
\centering
\caption{Full \textbf{MBPP} task results for \textbf{1B-parameter} models across various generation settings.}
\resizebox{\textwidth}{!}{%
\begin{tabular}{l l c l c c c c c}
\toprule
\textbf{Model} & \textbf{Generation Setting} & \textbf{PASS@1} & \shortstack{\textbf{Diffusion/Speculation} \\ \textbf{Sampling Strategy}} & \shortstack{\textbf{Acceptance} \\ \textbf{Rate (\%)}} & \shortstack{\textbf{Decoder} \\ \textbf{NFEs}} & \shortstack{\textbf{Global} \\ \textbf{NFEs}} & \shortstack{\textbf{Memory} \\ \textbf{Bandwidth} \\ \textbf{(GB)}} & \shortstack{\textbf{Memory} \\ \textbf{Decrease} \\ \textbf{vs. BLT (\%)}} \\
\midrule
\multirow{4}{*}{BLT 1B} & BLT (AR) & 14.00 & --- & --- & 256 & 220 & 654.61 & --- \\
\cmidrule(lr){2-9}
& \multirow{3}{*}{\shortstack[l]{BLT-S \\ (AR+self-speculation)}} & \multirow{3}{*}{14.00} & $k=4$ & 98.02 & 261 & 105 & 358.79 & 45.19 \\
& & & $k=8$ & 94.54 & 270 & 61 & 246.50 & 62.34 \\
& & & $k=16$ & 82.43 & 309 & 38 & 198.05 & 69.75 \\
\cmidrule(lr){1-9}
\multirow{8}{*}{BLT-D-4 1B}
    & \multirow{4}{*}{\shortstack[l]{BLT-D \\ (diffusion only)}} & 9.60 & Confidence-based, $\alpha=0.5$ & --- & 79 & 64 & 191.88 & 70.69 \\
    & & 12.60 & Confidence-based, $\alpha=0.7$ & --- & 92 & 64 & 195.97 & 70.06 \\
    & & 12.40 & EB sampling, $\gamma=0.8$ & --- & 110 & 64 & 201.84 & 69.17 \\
    & & 12.00 & EB sampling, $\gamma=1.0$ & --- & 105 & 64 & 200.09 & 69.43 \\
    \cmidrule(lr){2-9}
    & \multirow{4}{*}{\shortstack[l]{BLT-DV \\ (diffusion+verification)}} & \multirow{4}{*}{13.80} & Confidence-based, $\alpha=0.3$ & 96.42 & 109 & 106 & 312.00 & 52.34 \\
    & & & EB sampling, $\gamma=1.5$ & 97.30 & 132 & 105 & 317.04 & 51.57 \\
    & & & EB sampling, $\gamma=2.0$ & 97.10 & 126 & 105 & 315.65 & 51.78 \\
    & & & one step & 96.18 & 106 & 106 & 311.33 & 52.44 \\
\cmidrule(lr){1-9}
\multirow{8}{*}{BLT-D-8 1B}
    & \multirow{4}{*}{\shortstack[l]{BLT-D \\ (diffusion only)}} & 6.40 & Confidence-based, $\alpha=0.5$ & --- & 50 & 32 & 99.46 & 84.81 \\
    & & 8.60 & Confidence-based, $\alpha=0.7$ & --- & 64 & 32 & 103.89 & 84.13 \\
    & & 7.60 & EB sampling, $\gamma=0.8$ & --- & 92 & 32 & 112.74 & 82.78 \\
    & & 7.60 & EB sampling, $\gamma=1.0$ & --- & 86 & 32 & 110.91 & 83.06 \\
    \cmidrule(lr){2-9}
    & \multirow{4}{*}{\shortstack[l]{BLT-DV \\ (diffusion+verification)}} & \multirow{4}{*}{10.80} & Confidence-based, $\alpha=0.3$ & 91.56 & 68 & 62 & 184.69 & 71.79 \\
    & & & EB sampling, $\gamma=1.5$ & 93.79 & 103 & 61 & 192.32 & 70.62 \\
    & & & EB sampling, $\gamma=2.0$ & 94.00 & 95 & 61 & 189.54 & 71.05 \\
    & & & one step & 90.60 & 63 & 63 & 184.50 & 71.82 \\
\cmidrule(lr){1-9}
\multirow{8}{*}{BLT-D-16 1B}
    & \multirow{4}{*}{\shortstack[l]{BLT-D \\ (diffusion only)}} & 5.60 & Confidence-based, $\alpha=0.5$ & --- & 40 & 16 & 54.48 & 91.68 \\
    & & 8.20 & Confidence-based, $\alpha=0.7$ & --- & 56 & 16 & 59.66 & 90.89 \\
    & & 8.00 & EB sampling, $\gamma=0.8$ & --- & 89 & 16 & 69.91 & 89.32 \\
    & & 8.20 & EB sampling, $\gamma=1.0$ & --- & 82 & 16 & 67.74 & 89.65 \\
    \cmidrule(lr){2-9}
    & \multirow{4}{*}{\shortstack[l]{BLT-DV \\ (diffusion+verification)}} & \multirow{4}{*}{9.20} & Confidence-based, $\alpha=0.3$ & 79.32 & 53 & 38 & 117.97 & 81.98 \\
    & & & EB sampling, $\gamma=1.5$ & 86.31 & 100 & 35 & 125.17 & 80.88 \\
    & & & EB sampling, $\gamma=2.0$ & 85.85 & 91 & 36 & 122.99 & 81.21 \\
    & & & one step & 75.34 & 40 & 40 & 119.00 & 81.82 \\
\bottomrule
\end{tabular}
}
\label{tab:MBPP_results_1b}
\end{table*}

\newpage
\section{All 3B Model Results}
\label{app:all_3b_results}

In this section, we present the results of a larger sweep over inference hyperparameters for our 3B BLT-D, BLT-DV, and BLT-S models on the generation tasks. For BLT-D, we evaluate confidence-based unmasking with thresholds $\alpha \in \{0.5, 0.7\}$, as well as EB sampling with thresholds $\gamma \in \{0.8, 1.0\}$. For BLT-DV, we use more permissive settings that unmask more bytes per step; i.e., we \emph{decrease} $\alpha$ for confidence-based unmasking or \emph{increase} $\gamma$ for EB sampling. Specifically, we test $\alpha = 0.3$, $\gamma \in \{1.5, 2.0\}$, and one-step diffusion that unmasks all byte positions at once. For BLT-S, we use speculation windows $k \in \{4, 8, 16\}$. For BLT-S and BLT-DV, we also report the verification acceptance rate, defined as the fraction of drafted bytes that are accepted after verification. \Cref{tab:fr_en_results_3b}, \Cref{tab:de_en_results_3b}, \Cref{tab:human_eval_results_3b}, and \Cref{tab:MBPP_results_3b} report results on French-to-English translation, German-to-English translation, HumanEval, and MBPP, respectively.

\begin{table*}[htbp]
\centering
\caption{Full \textbf{French-to-English} translation results for \textbf{3B-parameter} models across various generation settings.}
\resizebox{\textwidth}{!}{%
\begin{tabular}{l l c l c c c c c}
\toprule
\textbf{Model} & \textbf{Generation Setting} & \textbf{BLEU} & \shortstack{\textbf{Diffusion/Speculation} \\ \textbf{Sampling Strategy}} & \shortstack{\textbf{Acceptance} \\ \textbf{Rate (\%)}} & \shortstack{\textbf{Decoder} \\ \textbf{NFEs}} & \shortstack{\textbf{Global} \\ \textbf{NFEs}} & \shortstack{\textbf{Memory} \\ \textbf{Bandwidth} \\ \textbf{(GB)}} & \shortstack{\textbf{Memory} \\ \textbf{Decrease} \\ \textbf{vs. BLT (\%)}} \\
\midrule
\multirow{4}{*}{BLT 3B} & BLT (AR) & 40.72 & --- & --- & 512 & 308 & 1920.99 & --- \\
\cmidrule(lr){2-9}
& \multirow{3}{*}{\shortstack[l]{BLT-S \\ (AR+self-speculation)}} & \multirow{3}{*}{40.72} & $k=4$ & 94.93 & 534 & 215 & 1395.99 & 27.33 \\
& & & $k=8$ & 87.16 & 580 & 130 & 928.73 & 51.65 \\
& & & $k=16$ & 69.93 & 724 & 87 & 727.17 & 62.15 \\
\cmidrule(lr){1-9}
\multirow{8}{*}{BLT-D-4 3B}
    & \multirow{4}{*}{\shortstack[l]{BLT-D \\ (diffusion only)}} & 37.75 & Confidence-based, $\alpha=0.5$ & --- & 185 & 128 & 787.36 & 59.01 \\
    & & 38.09 & Confidence-based, $\alpha=0.7$ & --- & 216 & 128 & 797.58 & 58.48 \\
    & & 37.79 & EB sampling, $\gamma=0.8$ & --- & 261 & 128 & 811.75 & 57.74 \\
    & & 37.83 & EB sampling, $\gamma=1.0$ & --- & 250 & 128 & 808.18 & 57.93 \\
    \cmidrule(lr){2-9}
    & \multirow{4}{*}{\shortstack[l]{BLT-DV \\ (diffusion+verification)}} & \multirow{4}{*}{38.89} & Confidence-based, $\alpha=0.3$ & 94.37 & 236 & 215 & 1300.92 & 32.28 \\
    & & & EB sampling, $\gamma=1.5$ & 95.37 & 290 & 213 & 1308.01 & 31.91 \\
    & & & EB sampling, $\gamma=2.0$ & 95.32 & 277 & 213 & 1304.23 & 32.11 \\
    & & & one step & 93.12 & 217 & 217 & 1307.60 & 31.93 \\
\cmidrule(lr){1-9}
\multirow{8}{*}{BLT-D-8 3B}
    & \multirow{4}{*}{\shortstack[l]{BLT-D \\ (diffusion only)}} & 35.94 & Confidence-based, $\alpha=0.5$ & --- & 143 & 64 & 409.85 & 78.66 \\
    & & 37.09 & Confidence-based, $\alpha=0.7$ & --- & 179 & 64 & 421.51 & 78.06 \\
    & & 36.54 & EB sampling, $\gamma=0.8$ & --- & 249 & 64 & 443.89 & 76.89 \\
    & & 36.60 & EB sampling, $\gamma=1.0$ & --- & 235 & 64 & 439.56 & 77.12 \\
    \cmidrule(lr){2-9}
    & \multirow{4}{*}{\shortstack[l]{BLT-DV \\ (diffusion+verification)}} & \multirow{4}{*}{38.66} & Confidence-based, $\alpha=0.3$ & 86.25 & 166 & 130 & 797.43 & 58.49 \\
    & & & EB sampling, $\gamma=1.5$ & 89.34 & 251 & 126 & 802.07 & 58.25 \\
    & & & EB sampling, $\gamma=2.0$ & 88.79 & 236 & 127 & 801.11 & 58.30 \\
    & & & one step & 84.63 & 133 & 133 & 799.94 & 58.36 \\
\cmidrule(lr){1-9}
\multirow{8}{*}{BLT-D-16 3B}
    & \multirow{4}{*}{\shortstack[l]{BLT-D \\ (diffusion only)}} & 31.64 & Confidence-based, $\alpha=0.5$ & --- & 123 & 32 & 221.58 & 88.47 \\
    & & 34.05 & Confidence-based, $\alpha=0.7$ & --- & 162 & 32 & 233.87 & 87.83 \\
    & & 33.75 & EB sampling, $\gamma=0.8$ & --- & 242 & 32 & 259.55 & 86.49 \\
    & & 33.69 & EB sampling, $\gamma=1.0$ & --- & 229 & 32 & 255.41 & 86.70 \\
    \cmidrule(lr){2-9}
    & \multirow{4}{*}{\shortstack[l]{BLT-DV \\ (diffusion+verification)}} & \multirow{4}{*}{35.23} & Confidence-based, $\alpha=0.3$ & 67.22 & 179 & 89 & 568.61 & 70.40 \\
    & & & EB sampling, $\gamma=1.5$ & 75.09 & 293 & 80 & 553.65 & 71.18 \\
    & & & EB sampling, $\gamma=2.0$ & 74.77 & 271 & 81 & 548.64 & 71.44 \\
    & & & one step & 60.33 & 99 & 99 & 598.66 & 68.84 \\
\bottomrule
\end{tabular}
}
\label{tab:fr_en_results_3b}
\end{table*}

\begin{table*}[htbp]
\centering
\caption{Full \textbf{German-to-English} translation results for \textbf{3B-parameter} models across various generation settings.}
\resizebox{\textwidth}{!}{%
\begin{tabular}{l l c l c c c c c}
\toprule
\textbf{Model} & \textbf{Generation Setting} & \textbf{BLEU} & \shortstack{\textbf{Diffusion/Speculation} \\ \textbf{Sampling Strategy}} & \shortstack{\textbf{Acceptance} \\ \textbf{Rate (\%)}} & \shortstack{\textbf{Decoder} \\ \textbf{NFEs}} & \shortstack{\textbf{Global} \\ \textbf{NFEs}} & \shortstack{\textbf{Memory} \\ \textbf{Bandwidth} \\ \textbf{(GB)}} & \shortstack{\textbf{Memory} \\ \textbf{Decrease} \\ \textbf{vs. BLT (\%)}} \\
\midrule
\multirow{4}{*}{BLT 3B} & BLT (AR) & 38.82 & --- & --- & 512 & 283 & 1776.54 & --- \\
\cmidrule(lr){2-9}
& \multirow{3}{*}{\shortstack[l]{BLT-S \\ (AR+self-speculation)}} & \multirow{3}{*}{38.82} & $k=4$ & 95.67 & 531 & 214 & 1387.43 & 21.90 \\
& & & $k=8$ & 87.82 & 576 & 129 & 922.56 & 48.07 \\
& & & $k=16$ & 70.23 & 721 & 86 & 724.53 & 59.22 \\
\cmidrule(lr){1-9}
\multirow{8}{*}{BLT-D-4 3B}
    & \multirow{4}{*}{\shortstack[l]{BLT-D \\ (diffusion only)}} & 35.74 & Confidence-based, $\alpha=0.5$ & --- & 186 & 128 & 787.95 & 55.65 \\
    & & 36.29 & Confidence-based, $\alpha=0.7$ & --- & 214 & 128 & 796.70 & 55.15 \\
    & & 36.48 & EB sampling, $\gamma=0.8$ & --- & 247 & 128 & 807.51 & 54.55 \\
    & & 36.53 & EB sampling, $\gamma=1.0$ & --- & 237 & 128 & 804.01 & 54.74 \\
    \cmidrule(lr){2-9}
    & \multirow{4}{*}{\shortstack[l]{BLT-DV \\ (diffusion+verification)}} & \multirow{4}{*}{37.46} & Confidence-based, $\alpha=0.3$ & 94.35 & 235 & 215 & 1300.96 & 26.77 \\
    & & & EB sampling, $\gamma=1.5$ & 95.08 & 279 & 214 & 1307.33 & 26.41 \\
    & & & EB sampling, $\gamma=2.0$ & 94.99 & 268 & 214 & 1304.90 & 26.55 \\
    & & & one step & 94.34 & 215 & 215 & 1294.63 & 27.13 \\
\cmidrule(lr){1-9}
\multirow{8}{*}{BLT-D-8 3B}
    & \multirow{4}{*}{\shortstack[l]{BLT-D \\ (diffusion only)}} & 33.83 & Confidence-based, $\alpha=0.5$ & --- & 146 & 64 & 411.05 & 76.86 \\
    & & 35.29 & Confidence-based, $\alpha=0.7$ & --- & 180 & 64 & 421.74 & 76.26 \\
    & & 35.41 & EB sampling, $\gamma=0.8$ & --- & 230 & 64 & 437.95 & 75.35 \\
    & & 35.46 & EB sampling, $\gamma=1.0$ & --- & 217 & 64 & 433.80 & 75.58 \\
    \cmidrule(lr){2-9}
    & \multirow{4}{*}{\shortstack[l]{BLT-DV \\ (diffusion+verification)}} & \multirow{4}{*}{37.11} & Confidence-based, $\alpha=0.3$ & 84.56 & 183 & 133 & 817.54 & 53.98 \\
    & & & EB sampling, $\gamma=1.5$ & 86.52 & 264 & 130 & 827.75 & 53.41 \\
    & & & EB sampling, $\gamma=2.0$ & 86.37 & 249 & 130 & 824.42 & 53.59 \\
    & & & one step & 81.62 & 137 & 137 & 826.79 & 53.46 \\
\cmidrule(lr){1-9}
\multirow{8}{*}{BLT-D-16 3B}
    & \multirow{4}{*}{\shortstack[l]{BLT-D \\ (diffusion only)}} & 30.30 & Confidence-based, $\alpha=0.5$ & --- & 109 & 32 & 217.12 & 87.78 \\
    & & 32.62 & Confidence-based, $\alpha=0.7$ & --- & 136 & 32 & 225.81 & 87.29 \\
    & & 32.56 & EB sampling, $\gamma=0.8$ & --- & 206 & 32 & 247.99 & 86.04 \\
    & & 32.48 & EB sampling, $\gamma=1.0$ & --- & 191 & 32 & 243.35 & 86.30 \\
    \cmidrule(lr){2-9}
    & \multirow{4}{*}{\shortstack[l]{BLT-DV \\ (diffusion+verification)}} & \multirow{4}{*}{34.52} & Confidence-based, $\alpha=0.3$ & 74.49 & 161 & 83 & 524.04 & 70.50 \\
    & & & EB sampling, $\gamma=1.5$ & 77.89 & 263 & 79 & 534.83 & 69.89 \\
    & & & EB sampling, $\gamma=2.0$ & 77.66 & 245 & 79 & 530.54 & 70.14 \\
    & & & one step & 70.44 & 88 & 88 & 529.76 & 70.18 \\
\bottomrule
\end{tabular}
}
\label{tab:de_en_results_3b}
\end{table*}

\begin{table*}[htbp]
\centering
\caption{Full \textbf{HumanEval} task results for \textbf{3B-parameter} models across various generation settings.}
\resizebox{\textwidth}{!}{%
\begin{tabular}{l l c l c c c c c}
\toprule
\textbf{Model} & \textbf{Generation Setting} & \textbf{PASS@1} & \shortstack{\textbf{Diffusion/Speculation} \\ \textbf{Sampling Strategy}} & \shortstack{\textbf{Acceptance} \\ \textbf{Rate (\%)}} & \shortstack{\textbf{Decoder} \\ \textbf{NFEs}} & \shortstack{\textbf{Global} \\ \textbf{NFEs}} & \shortstack{\textbf{Memory} \\ \textbf{Bandwidth} \\ \textbf{(GB)}} & \shortstack{\textbf{Memory} \\ \textbf{Decrease} \\ \textbf{vs. BLT (\%)}} \\
\midrule
\multirow{4}{*}{BLT 3B} & BLT (AR) & 22.56 & --- & --- & 512 & 250 & 1590.45 & --- \\
\cmidrule(lr){2-9}
& \multirow{3}{*}{\shortstack[l]{BLT-S \\ (AR+self-speculation)}} & \multirow{3}{*}{22.56} & $k=4$ & 98.68 & 518 & 208 & 1353.39 & 14.91 \\
& & & $k=8$ & 95.96 & 532 & 120 & 853.11 & 46.36 \\
& & & $k=16$ & 88.01 & 581 & 70 & 585.81 & 63.17 \\
\cmidrule(lr){1-9}
\multirow{8}{*}{BLT-D-4 3B}
    & \multirow{4}{*}{\shortstack[l]{BLT-D \\ (diffusion only)}} & 17.07 & Confidence-based, $\alpha=0.5$ & --- & 144 & 128 & 774.41 & 51.31 \\
    & & 18.90 & Confidence-based, $\alpha=0.7$ & --- & 159 & 128 & 779.20 & 51.01 \\
    & & 18.90 & EB sampling, $\gamma=0.8$ & --- & 188 & 128 & 788.50 & 50.42 \\
    & & 18.90 & EB sampling, $\gamma=1.0$ & --- & 180 & 128 & 785.82 & 50.59 \\
    \cmidrule(lr){2-9}
    & \multirow{4}{*}{\shortstack[l]{BLT-DV \\ (diffusion+verification)}} & \multirow{4}{*}{18.90} & Confidence-based, $\alpha=0.3$ & 97.97 & 214 & 208 & 1257.30 & 20.95 \\
    & & & EB sampling, $\gamma=1.5$ & 98.29 & 239 & 208 & 1262.78 & 20.60 \\
    & & & EB sampling, $\gamma=2.0$ & 98.26 & 232 & 208 & 1260.65 & 20.74 \\
    & & & one step & 97.74 & 209 & 209 & 1258.33 & 20.88 \\
\cmidrule(lr){1-9}
\multirow{8}{*}{BLT-D-8 3B}
    & \multirow{4}{*}{\shortstack[l]{BLT-D \\ (diffusion only)}} & 10.37 & Confidence-based, $\alpha=0.5$ & --- & 88 & 64 & 392.51 & 75.32 \\
    & & 15.85 & Confidence-based, $\alpha=0.7$ & --- & 106 & 64 & 398.04 & 74.97 \\
    & & 15.24 & EB sampling, $\gamma=0.8$ & --- & 152 & 64 & 412.75 & 74.05 \\
    & & 15.24 & EB sampling, $\gamma=1.0$ & --- & 142 & 64 & 409.53 & 74.25 \\
    \cmidrule(lr){2-9}
    & \multirow{4}{*}{\shortstack[l]{BLT-DV \\ (diffusion+verification)}} & \multirow{4}{*}{16.46} & Confidence-based, $\alpha=0.3$ & 94.51 & 130 & 120 & 728.09 & 54.22 \\
    & & & EB sampling, $\gamma=1.5$ & 96.04 & 176 & 118 & 733.00 & 53.91 \\
    & & & EB sampling, $\gamma=2.0$ & 95.91 & 165 & 119 & 730.33 & 54.08 \\
    & & & one step & 93.63 & 121 & 121 & 731.02 & 54.04 \\
\cmidrule(lr){1-9}
\multirow{8}{*}{BLT-D-16 3B}
    & \multirow{4}{*}{\shortstack[l]{BLT-D \\ (diffusion only)}} & 8.54 & Confidence-based, $\alpha=0.5$ & --- & 62 & 32 & 201.98 & 87.30 \\
    & & 9.76 & Confidence-based, $\alpha=0.7$ & --- & 84 & 32 & 208.94 & 86.86 \\
    & & 11.59 & EB sampling, $\gamma=0.8$ & --- & 143 & 32 & 227.97 & 85.67 \\
    & & 10.98 & EB sampling, $\gamma=1.0$ & --- & 133 & 32 & 224.65 & 85.88 \\
    \cmidrule(lr){2-9}
    & \multirow{4}{*}{\shortstack[l]{BLT-DV \\ (diffusion+verification)}} & \multirow{4}{*}{14.02} & Confidence-based, $\alpha=0.3$ & 83.61 & 94 & 72 & 445.22 & 72.01 \\
    & & & EB sampling, $\gamma=1.5$ & 87.77 & 172 & 69 & 449.96 & 71.71 \\
    & & & EB sampling, $\gamma=2.0$ & 87.61 & 155 & 69 & 445.39 & 72.00 \\
    & & & one step & 80.68 & 75 & 75 & 453.15 & 71.51 \\
\bottomrule
\end{tabular}
}
\label{tab:human_eval_results_3b}
\end{table*}

\begin{table*}[htbp]
\centering
\caption{Full \textbf{MBPP} task results for \textbf{3B-parameter} models across various generation settings.}
\resizebox{\textwidth}{!}{%
\begin{tabular}{l l c l c c c c c}
\toprule
\textbf{Model} & \textbf{Generation Setting} & \textbf{PASS@1} & \shortstack{\textbf{Diffusion/Speculation} \\ \textbf{Sampling Strategy}} & \shortstack{\textbf{Acceptance} \\ \textbf{Rate (\%)}} & \shortstack{\textbf{Decoder} \\ \textbf{NFEs}} & \shortstack{\textbf{Global} \\ \textbf{NFEs}} & \shortstack{\textbf{Memory} \\ \textbf{Bandwidth} \\ \textbf{(GB)}} & \shortstack{\textbf{Memory} \\ \textbf{Decrease} \\ \textbf{vs. BLT (\%)}} \\
\midrule
\multirow{4}{*}{BLT 3B} & BLT (AR) & 29.60 & --- & --- & 256 & 222 & 1349.34 & --- \\
\cmidrule(lr){2-9}
& \multirow{3}{*}{\shortstack[l]{BLT-S \\ (AR+self-speculation)}} & \multirow{3}{*}{29.60} & $k=4$ & 98.21 & 260 & 105 & 685.35 & 49.21 \\
& & & $k=8$ & 94.84 & 269 & 61 & 436.89 & 67.62 \\
& & & $k=16$ & 84.41 & 302 & 37 & 310.63 & 76.98 \\
\cmidrule(lr){1-9}
\multirow{7}{*}{BLT-D-4 3B}
    & \multirow{4}{*}{\shortstack[l]{BLT-D \\ (diffusion only)}} & 24.60 & Confidence-based, $\alpha=0.5$ & --- & 76 & 64 & 388.71 & 71.19 \\
    & & 26.00 & Confidence-based, $\alpha=0.7$ & --- & 89 & 64 & 392.59 & 70.90 \\
    & & 25.80 & EB sampling, $\gamma=0.8$ & --- & 107 & 64 & 398.39 & 70.48 \\
    & & 25.80 & EB sampling, $\gamma=1.0$ & --- & 101 & 64 & 396.72 & 70.60 \\
    \cmidrule(lr){2-9}
    & \multirow{3}{*}{\shortstack[l]{BLT-DV \\ (diffusion+verification)}} & \multirow{3}{*}{27.20} & EB sampling, $\gamma=1.5$ & 97.94 & 129 & 105 & 639.19 & 52.63 \\
    & & & EB sampling, $\gamma=2.0$ & 97.86 & 123 & 105 & 637.77 & 52.74 \\
    & & & one step & 96.98 & 105 & 105 & 635.95 & 52.87 \\
\cmidrule(lr){1-9}
\multirow{8}{*}{BLT-D-8 3B}
    & \multirow{4}{*}{\shortstack[l]{BLT-D \\ (diffusion only)}} & 18.40 & Confidence-based, $\alpha=0.5$ & --- & 49 & 32 & 197.75 & 85.34 \\
    & & 20.80 & Confidence-based, $\alpha=0.7$ & --- & 63 & 32 & 202.22 & 85.01 \\
    & & 23.20 & EB sampling, $\gamma=0.8$ & --- & 88 & 32 & 210.29 & 84.42 \\
    & & 22.40 & EB sampling, $\gamma=1.0$ & --- & 82 & 32 & 208.37 & 84.56 \\
    \cmidrule(lr){2-9}
    & \multirow{4}{*}{\shortstack[l]{BLT-DV \\ (diffusion+verification)}} & \multirow{4}{*}{27.00} & Confidence-based, $\alpha=0.3$ & 92.68 & 67 & 61 & 373.68 & 72.31 \\
    & & & EB sampling, $\gamma=1.5$ & 94.85 & 99 & 60 & 376.71 & 72.08 \\
    & & & EB sampling, $\gamma=2.0$ & 94.65 & 92 & 60 & 375.13 & 72.20 \\
    & & & one step & 91.87 & 62 & 62 & 374.64 & 72.24 \\
\cmidrule(lr){1-9}
\multirow{8}{*}{BLT-D-16 3B}
    & \multirow{4}{*}{\shortstack[l]{BLT-D \\ (diffusion only)}} & 10.60 & Confidence-based, $\alpha=0.5$ & --- & 39 & 16 & 103.64 & 92.32 \\
    & & 15.80 & Confidence-based, $\alpha=0.7$ & --- & 55 & 16 & 108.78 & 91.94 \\
    & & 15.60 & EB sampling, $\gamma=0.8$ & --- & 88 & 16 & 119.48 & 91.15 \\
    & & 15.60 & EB sampling, $\gamma=1.0$ & --- & 81 & 16 & 117.21 & 91.31 \\
    \cmidrule(lr){2-9}
    & \multirow{4}{*}{\shortstack[l]{BLT-DV \\ (diffusion+verification)}} & \multirow{4}{*}{19.00} & Confidence-based, $\alpha=0.3$ & 74.77 & 56 & 41 & 251.34 & 81.37 \\
    & & & EB sampling, $\gamma=1.5$ & 81.05 & 107 & 37 & 250.43 & 81.44 \\
    & & & EB sampling, $\gamma=2.0$ & 80.38 & 97 & 38 & 249.14 & 81.54 \\
    & & & one step & 71.79 & 42 & 42 & 255.40 & 81.07 \\
\bottomrule
\end{tabular}
}
\label{tab:MBPP_results_3b}
\end{table*}

\end{document}